\def\year{2015}
\documentclass[letterpaper]{article}
\usepackage{color}
\usepackage{aaai}
\usepackage{algorithm}
\usepackage{algpseudocode}
\usepackage{amsmath}
\usepackage{amsfonts}
\usepackage{amssymb}
\usepackage{mathtools}
\usepackage{times}
\usepackage{helvet}
\usepackage{courier}
\usepackage{url}

\usepackage{tikz}
\usepackage{tikz-qtree}
\usetikzlibrary{backgrounds,arrows,calc,positioning,shapes, decorations.markings}

\DeclareMathOperator*{\argmax}{\arg\!\max}

\usepackage[normalem]{ulem}

\DeclarePairedDelimiter\norm{\Vert}{\Vert}

\DeclarePairedDelimiter\floor{\lfloor}{\rfloor}

\newcommand{\knownness}{{\kappa}}
\newcommand{\kdtree}{{\tau}}
\newcommand{\hrect}{{\mathcal{H}}}
\newcommand{\cardinality}[1]{{\left\vert{#1}\right\vert}}
\newcommand{\hrectSize}[1]{\cardinality{#1}}
\newcommand{\leaf}[2]{{\mathrm{leaf}(#1,#2)}}
\newcommand{\leafSpace}[1]{{\mathrm{space}(#1)}}
\newcommand{\leafPoints}[1]{{\mathrm{points}(#1)}}

\newcommand{\norminf}[1]{\norm{#1}_\infty}

\nocopyright

\begin{document}
%
\title{Online Reinforcement Learning for Real-Time Exploration in Continuous
       State and Action Markov Decision Processes}
\author{
        Hofer Ludovic \and Gimbert Hugo\\
        Laboratoire Bordelais de Recherche en Informatique\\
        351 Cours de Libération\\
        33405 Talence\\
        France             
}
\maketitle

\begin{abstract}
\begin{quote}
This paper presents a new method to learn online policies in continuous state,
continuous action, model-free Markov decision processes, with two properties
that are crucial for practical applications.  First, the policies are
implementable with a very low computational cost: once the policy is computed,
the action corresponding to a given state is obtained in logarithmic time with
respect to the number of samples used.  Second, our method is versatile: it does
not rely on any a priori knowledge of the structure of optimal policies.  We
build upon the Fitted Q-iteration algorithm which represents the $Q$-value as the
average of several regression trees. Our algorithm, the Fitted Policy Forest
algorithm (FPF), computes a regression forest representing the Q-value and
transforms it into a single tree representing the policy, while keeping control on
the size of the policy using resampling and leaf merging. We introduce an
adaptation of Multi-Resolution Exploration (MRE) which is particularly suited to
FPF.  We assess the performance of FPF on three classical benchmarks for
reinforcement learning: the "Inverted Pendulum", the "Double Integrator" and
"Car on the Hill" and show that FPF equals or outperforms other algorithms,
although these algorithms rely on the use of particular representations of the
policies, especially chosen in order to fit each of the three problems. Finally,
we exhibit that the combination of FPF and MRE allows to find nearly optimal
solutions in problems where $\epsilon$-greedy approaches would fail.
\end{quote}
\end{abstract}

\section{\label{sec:Introduction}Introduction}
\begin{figure}
  \centering
  \includegraphics[width=.8\linewidth]{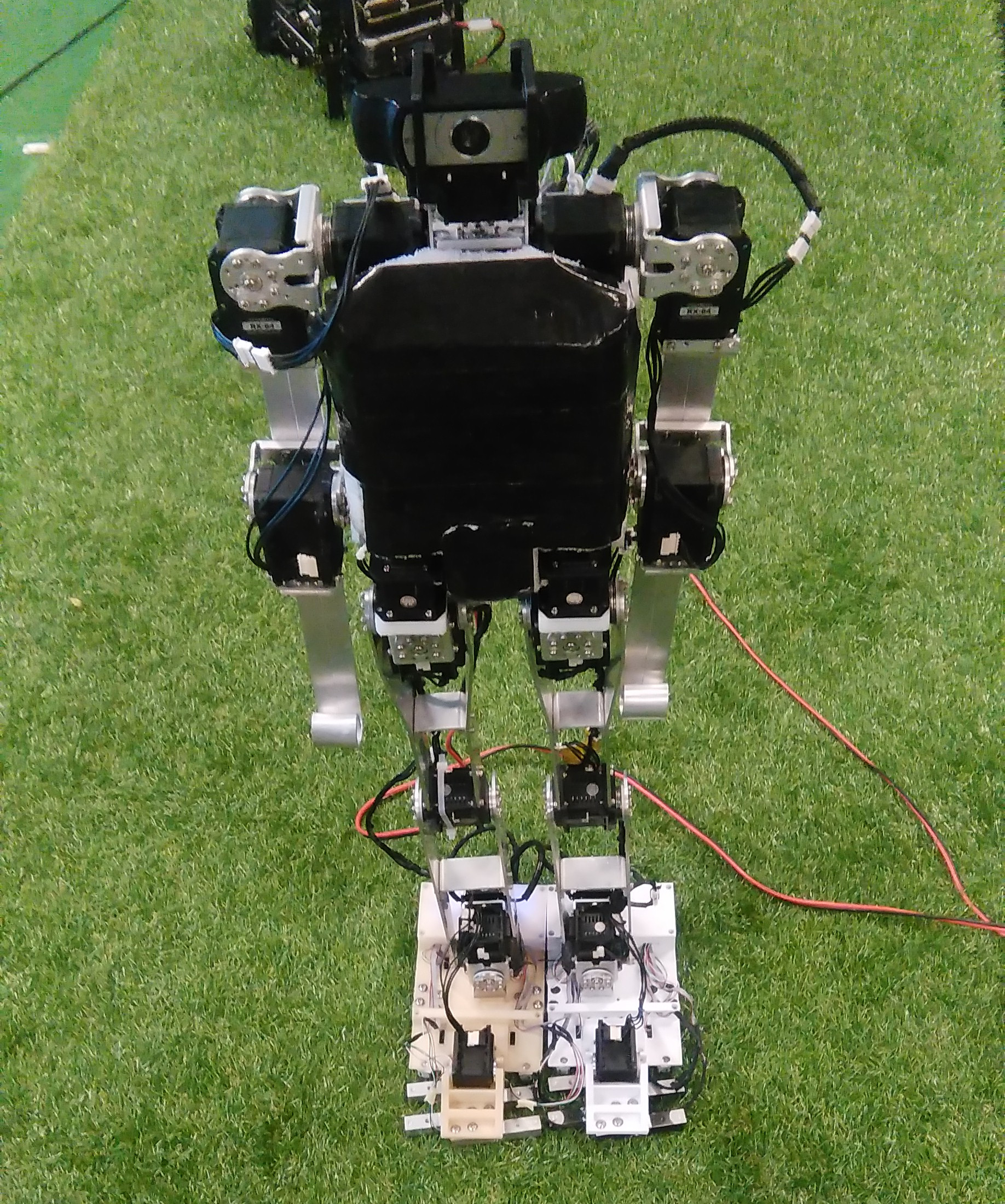}
  \caption{\label{fig:Grosban}The Grosban robot}
\end{figure}

The initial motivation for the research presented in this paper
is the optimization of closed-loop control of humanoid robots,
autonomously playing soccer at the annual Robocup
competition~\footnote{http://wiki.robocup.org/wiki/Humanoid\_League}. We
specifically target to learn behaviors on the Grosban robot, presented in
Figure~\ref{fig:Grosban}.
This requires the computation of policies in Markov decision processes where
1) the state space is continous, 2) the action space is continous, 3) the 
transition function is not known. Additionally, in order to provide real-time
closed-loop control, the policy should allow to retrieve a nearly optimal-action
at a low computational-cost.
We consider that the transition function is not known, because with small and
low-cost humanoid robots, the lack of accuracy on sensors and effectors makes
the system behavior difficult to predict.

More generally, the control of physical systems naturally leads to models with
continous-action spaces, since one typically controls the position and
acceleration of an object or the torque sent to a joint. While policy gradients
methods have been used successfully to learn highly dynamical tasks such as
hitting a baseball with an anthropomorphic arm~\cite{Peters2008}, those
algorithms are not suited for learning on low-cost robots, because they need to
provide a motor primitive and to be able to estimate a gradient of the reward
with respect to the motor primitive parameters. While model-based
control is difficult to apply on such robots, hand-tuned open-loop behaviors
have proven to be very effective~\cite{Behnke2006}. Therefore, model-free
learning for CSA-MDP appears as a promising approach to learn such behaviors.

Since the transition and the reward functions are not known \emph{a priori},
sampling is necessary. While an efficient exploitation of the collected samples
is required, it is not sufficient. A smart exploration is necessary, because on
some problems, nearly-optimal strategies requires a succession of actions which
is very unlikely to occur when using uniformous random actions. On extreme cases,
it might even lead to situation where no reward is ever seen, because the
probability of reaching a state carrying a reward while following a random
policy is almost 0. This problem is known as the combinatory lock problem
and appears in discrete case in~\cite{Koenig1996} and in continuous problems
in~\cite{Li2009}.

For control problems where the action set is discrete and not too large, there
are already existing efficient algorithms to tackle the problem of producing an
efficient policy from the result of previous experiments.  Of course, these
algorithms can be used in the continous action space case, by discretization of
the action sets. However this naive approach often leads to computational costs
that are too high for practical applications, as stated in~\cite{Weinstein2014}.

The specificity of continuous action space has also been adressed with specific
methods and particularly encouraging empirical results have been obtained thanks
for example to the \emph{Binary Action Search} approach~\cite{Pazis2009}, see
also~\cite{Busoniu2010}.  These methods require to design
functional basis used to represent the $Q$-value function, which we prefer to
avoid in order to obtain versatile algorithms.

A recent major-breakthrough in the field of solving CSA-MDP is Symbolic Dynamic
Programming which allows to find exact solutions by using eXtended Algebraic
Decision Diagrams~\cite{Sanner2012}, see also~\cite{Zamani2012}. However, those
algorithms requires a model of the MDP and rely on several assumptions
concerning the shape of the transition function and the reward function.
Additionally, those methods are suited for a very close horizon and are
therefore not suited for our application.

While local planning allows to achieve outstanding control on high-dimensionnal
problems such as humanoid locomotion~\cite{Weinstein2013}, the computational
cost of online planning is a burden for real-time application. This is
particularly relevant in robotics, where processing units have to be light and
small in order to be embedded. Therefore, we aim at global planning, where the
policy is computed offline and then loaded on the robot.

Our own learning algorithms are based on the Fitted $Q$ Iteration
algorithm~\cite{Ernst2005} which represents the $Q$-value as the average of
several regression trees. We first present a method allowing to extract
approximately optimal continuous action from a $Q$-value forest. Then we
introduce a new algorithm, Fitted Policy Forest (FPF), which learn an
approximation of the policy function using regression forests. Such a
representation of the policy allows to retrieve a nearly optimal action at a
very low computational cost, therefore allowing to use it on embedded systems.

We use an exploration algorithm based on MRE~\cite{Nouri2009}, an optimistic
algorithm which represents the knownness of state and action couples using a
kd-tree~\cite{Preparata1985}.  Following the idea of extremely randomized
trees~\cite{Geurts2006}, we introduce randomness in the split, thus allowing to
grow a forest in order to increase the smoothness of the knownness
function. Moreover, by changing the update rule for the $Q$-value, we reduce the
attracting power of local maxima.

The viability of FPF is demonstrated by a performance comparison with the
results proposed in~\cite{Pazis2009} on three classical benchmark in
RL: \emph{Inverted Pendulum Stabilization}, \emph{Double Integrator} and
\emph{Car on the Hill}. Experimental results show that FPF drastically reduce the
computation time while improving performance. We further illustrate the gain
obtained by using our version of MRE on the \emph{Inverted Pendulum
Stabilization} problem, we finally present the results obtained on
the \emph{Inverted Pendulum Swing-Up}, using an underactuated angular
joint. This last experiment is run using Gazebo simulator in place of the
analytical model.

This paper is organized as follows: Section~\ref{sec:Background} introduces the
notations used for Markov decision processes and regression forests,
Section~\ref{sec:PreviousWork} presents the original version of
\emph{Fitted $Q$-Iteration} and other classical methods in batch mode RL with
continuous action space, Section~\ref{sec:QValue} proposes algorithms to extract
informations from regression forest, Section~\ref{sec:LearningPolicy} introduces
the core of the FPF algorithm. Section~\ref{sec:Exploration} presents the
exploration algorithm we used. The efficiency of FPF and MRE is demonstrated
through a series of experiments on classical RL benchmarks in
section~\ref{sec:Experimentations}, the meaning of the experimental results is
discussed in Section~\ref{sec:Discussion}.

\section{\label{sec:Background}Background}

\subsection{Markov-Decision Process}
A \emph{Markov-Decision Process}, or MDP for short, is a 5-tuple
$\langle S,A,R,T,\gamma \rangle$, where $S$ is a set of states, $A$ is a set of actions,
$R$ is a reward function ($R(s,a)$ denotes the expected reward when taking action
$a$ in state $s$), $T$ is the transition function ($T(s,a,s')$ denotes
the probability of reaching $s'$ from $s$ using $a$) and $\gamma \in [0,1[$ is a
discount factor.

A \emph{Deterministic Policy} is a mapping $\pi: S \mapsto A$, where $\pi(s)$
denotes the action choice in state $s$. Thereafter, by ``policy'', we
implicitely refer to deterministic policy. The $Q$-value of a couple $(s,a)$
under a policy $\pi$ with an horizon $H$ is denoted $Q^\pi_H(s,a)$ and is
defined as the expected cumulative and discounted reward by applying $a$ in
state $s$ and then choosing actions according to $\pi$:
$$Q^\pi_H(s,a) = R(s,a) + \gamma  \sum_{s'\in S}T(s,a,s') Q^\pi_{H-1}(s',\pi(s'))$$
We further abreviate $Q^\pi_\infty$ by $Q^\pi$ for short. The greedy policy with
respect to $Q$ is denoted $\pi_Q$ and always selects the action with the highest
$Q$-value; i.e.  $\pi_Q(s) = \argmax\limits_{a \in A} Q(s,a)$. Considering that
the action space is bounded to an interval, such a limit exists, although it is
not necessarily unique.

It is known that an optimal $Q$-value function exists~\cite{Puterman1994}:
$Q^* = \max\limits_\pi{Q^\pi}$. The optimal policy $\pi^*$ is greedy with respect
to $Q^*$: $\pi^* = \pi_{Q^*}$.

Given a complete and finite MDP, standard algorithms exists for finding the
optimal policy, including value iteration, policy iteration and linear
programming. However, if the transition function or the reward function are
unknown, it is necessary to use samples to learn an approximation of the
$Q$value denoted $\widehat{Q}$. If the state space or the action space
are continuous, it is also necessary to approximate the solution.

When solving offline a MDP while having no direct access to the transition
function, it is necessary to use a set of gathered samples. Samples are defined
as 4-tuples of the form: $\langle s,a,r,s' \rangle$ where $s$ is the starting
state, $a$ the action used, $r$ the reward received and $s'$ the successor
state.

\subsection{Regression Forests}
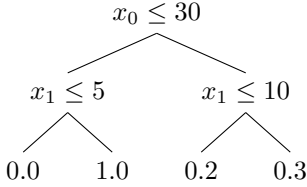
\begin{figure}
  \centering
  \begin{tikzpicture} [sibling distance=5mm]
    \Tree
        [.$x_0\leq30$
          [.$x_1\leq5$ $0.0$ $1.0$ ]
          [.$x_1\leq10$ $0.2$ $0.3$ ]
        ]
  \end{tikzpicture}
  \caption{\label{fig:simpleTree}A simple regression tree}
\end{figure}

A \emph{regression tree} is a representation of the approximation of a function
$f: X \mapsto Y$ where $X \in \mathbb{R}^k$ and $Y \in \mathbb{R}$. It has a
decision tree structure where every non-leaf node is a function mapping $X$ to
its children and every leaf is a basic function $\phi: X \mapsto Y$. A simple
regression tree with piecewise constant (PWC) approximation is presented in
Figure~\ref{fig:simpleTree}. Several algorithms exist to extract regression
trees from training set, for a complete introduction, refer to~\cite{Loh2011}.
Predicting the output $y$ from an entry $x$ requires to find the leaf
corresponding to $x$ and then to compute $\phi(x)$, with $\phi$ the basic function
found at the leaf corresponding to $x$. We will further refer to the value
predicted by a tree $t$ for input $x$ by $t(x)$ for short. While some algorithms
uses oblique split~\cite{Li2000}, the algorithms presented here are only valid
for orthogonal splits (splits of the form $x_i \leq v$). We will further note
$\mathrm{LC}(n)$ and $\mathrm{UC}(n)$ the lower and upper children of node $n$,
concerning $x_i \leq v$ and $x_i > v$ respectively.

If we define the space $X$ as a hyperrectangle $\hrect$, each leaf will
concern a different part of $\hrect$. We will further refer to the minimun
and maximum value of $\hrect$ along the dimension $i$ as $\hrect_{i,m}$
and $\hrect_{i,M}$ respectively. We define the size of a hyperrectangle
$\hrect$ by
$\hrectSize{\hrect} = \prod\limits_{i = 1}^{\dim{X}}\hrect_{i,M} - \hrect_{i,m}$.
We use an abusive notation of the norm $\norm{\hrect}$ in place of
$\norm{\hrect_{i,M} -\hrect_{i,m}}$.

A \emph{regression forest} is a set of regression trees:
$F = \{t_1, \dots, t_M\}$. It has been exhibited in~\cite{Breiman1996} that using
multipe trees to represent the function leads to a more accurate prediction.
The value predicted by a forest $F$ for an input $x$ is
$F(x) = \sum\limits_{k=1}^M\frac{t_k(x)}{M}$.

\subsection{Kd-trees}
Kd-trees are a data structure which allows to store points of the same size
while providing an $O(\log (n))$ access~\cite{Preparata1985}. At each leaf of
the tree, there is one or several points and at each non-leaf node, there is an
orthogonal split. Let $X$ be the space on which the kd-tree $\kdtree$ is
defined, then for every $x \in X$, there exist a single path from the root of
the kd-tree to the leaf in which $x$ would fit. This leaf is denoted
$\leaf{\kdtree}{x}$ is defined on the space $X$. Each leaf $l$ contains a set
of points noted $\leafPoints{l}$ and concerns an hyperrectangle
$\hrect = \leafSpace{l}$.

\section{\label{sec:PreviousWork}Previous Work}
The use of regression forests to approximate the $Q$-value of a continuous MDP
has been introduced in~\cite{Ernst2005} under the name of
\emph{Fitted Q Iteration}. This algorithm uses an iterative procedure to build
$\widehat{Q_H}$, an approximation of the $Q$-value function at horizon $H$. It
builds a regression forest by using $\widehat{Q_{H-1}}$ and a set of 4-tuples
using the rules given at Equations~\ref{eq:FQI:input} and~\ref{eq:FQI:output}.

\begin{equation}
        x = (s,a)
\label{eq:FQI:input}
\end{equation}

\begin{equation}
        y = r + \max\limits_{a \in A}\widehat{Q_{H-1}}(s',a)
\label{eq:FQI:output}
\end{equation}

While this procedure yields very satisfying results when the action space is
discrete, the computational complexity of the $\max$ part in
equation~\ref{eq:FQI:output} when using regression forest makes it become
quickly inefficient. Therefore, in~\cite{Ernst2005}, action spaces are always
discretized to compute this equation, thus leading to an inappropriate action
set when optimal control requires a fine discretization.

\emph{Binary Action Search}, introduced in~\cite{Pazis2009} proposes a generical
approach allowing to avoid the computation of the $\max$ part in
equation~\ref{eq:FQI:output}. Results presented in~\cite{Pazis2009} show that
Binary Action Search strongly outperforms method with a finite number of actions
on two problems with rewards including a cost depending on the square of the
action used: \emph{Inverted Pendulum Stabilization} and \emph{Double
Integrator}. On the other hand, binary action search yields unsatisfying results
on \emph{Car on the Hill}, a problem with an optimal strategy known to be
``bang-bang'' (i.e. optimal strategy is only composed of two actions).

\section{\label{sec:QValue}Approximation of the $Q$-value forest}
In this part, we propose new methods to extract information from a regression forest while
choosing a trade-off between accuracy and computational cost.
First, we introduce the algorithm we use to grow regression forest. Then we
present an algorithm to project a regression tree on a given subspace.
Finally we propose a method allowing to average a whole regression forest by a
single regression tree whose number of leaf is bounded.

\subsection{Extra-Trees}
While several methods exists to build regression forests from a training
samples, our implementation is based on Extra-Trees~\cite{Geurts2006}.
This
algorithm produces satisfying approximation at a moderate computational cost.

The main characteristic of Extra-trees is that $k$ split dimensions are chosen
randomly, then for each chosen split dimension the position of the split is
picked randomly from an uniformous distribution from the minimal to the maximal
value of the dimension along the samples to split. Finally, only the best of the
$k$ random splits is used; the criteria used to rank the splits is the variance
gain brought by the split. The original training set is splitted until one of
the terminal condition is reached. The first terminal condition is that the
number of samples remaining is smaller than $n_\mathrm{min}$, where
$n_\mathrm{min}$ is a parameter allowing to control overfitting.
There are two other terminal conditions: if the inputs of the samples are
all identical or if the output value is constant.

\subsection{Improving Extra-trees}
We provide two improvements to Extra-trees, in order to remedy two problems.
First, due to the terminal conditions, large trees are grown for parts of the
space were the $Q$-value is almost constant because if the $Q$-value is not
strictly constant, the only terminal condition is that the number of samples is
lower than $n_\mathrm{min}$. We remedy this problem with the help of a new
parameter $V_\mathrm{min}$ which specifies the minimal variance
between prediction and measure necessary to allow splitting.
A naive implementation of Extra-Trees leads to a second problem:
it may generate nodes with very few samples, which paves the way to overfitting
and is bad for linear interpolation. Therefore, we changed
the choice of the split values. Instead of choosing it uniformly from the
minimum to the maximum of the samples, our algorithm choose it uniformly between
the $n_\mathrm{min}$-th smallest and highest values, which guarantees that each
node of the split tree contains at least $n_\mathrm{min}$ samples.

\subsection{Projection of a regression tree}
Let consider a tree $t: S \times A \mapsto \mathbb{R}$, we can define the
projection of the tree $t$ on the state $s$ as another tree
$\mathcal{P}(t,s) = t': A \mapsto \mathbb{R}$. Since $s$ is known, $t'$ does
not contain any split depending on $s$ value and therefore contains only splits
related to the action space. It is easy to create a hyperrectangle
$\hrect$ corresponding to state $s$.
$$\hrect(s) =
                 \begin{pmatrix}
                        s_1&s_1\\
                        \vdots&\vdots\\
                        s_{D_S}&s_{D_s}\\
                        \mathrm{min}(A_1)&\mathrm{max}(A_1)\\
                        \vdots&\vdots\\
                        \mathrm{min}(A_{D_A})&\mathrm{max}(A_{D_A})\\
                 \end{pmatrix}$$

The pseudo-code for tree projection is shown in
Algorithm~\ref{alg:treeProjection}.

\begin{algorithm}
  \begin{algorithmic}[1]
    \Function{projectTree}{$t$,$\hrect$}
      \State \Return projectNode(root($t$), $\hrect$)
    \EndFunction
    \Function{projectNode}{node,$\hrect$}
      \If{isLeaf(node)}
        \State \Return node
      \EndIf
      \State $d \gets \mathrm{splitDim(node)}$
      \State $v \gets \mathrm{splitVal(node)}$
      \If{$v > \hrect_{d,M}$}
        \State node $\gets$ projectTree(LC(node),$\hrect$)
      \ElsIf{$v \leq \hrect_{d,m}$}
        \State node $\gets$ projectTree(UC(node),$\hrect$)
      \Else
        \State LC(node) $\gets$ projectTree(LC(node),$\hrect$)
        \State UC(node) $\gets$ projectTree(UC(node),$\hrect$)
      \EndIf
      \State \Return node
    \EndFunction
  \end{algorithmic}
  \caption{\label{alg:treeProjection} The tree projection algorithm}
\end{algorithm}

\subsection{Weighted average of regression trees}
Let $t_1$ and $t_2$ be two regressions trees mapping $X$ to $Y$, weight
respectively by $w_1$ and $w_2$, we define the weighted average of the trees
as a tree $t' = \mu(t_1,t_2,w_1,w_2)$ such as:
$$\forall x \in X, t'(x) = \frac{t_1(x)w_1 + t_2(x)w_2}{w_1 + w_2}$$
A simple scheme for computing $t'$ would be to root a replicate of $t_2$ at
each leaf of $t_1$. However this would lead to an overgrown tree containing
various unreachable nodes. As example, a split with the predicate $x_1 \leq 3$
could perfectly appear on the lower child of another node whose predicate is
$x_1 \leq 2$.

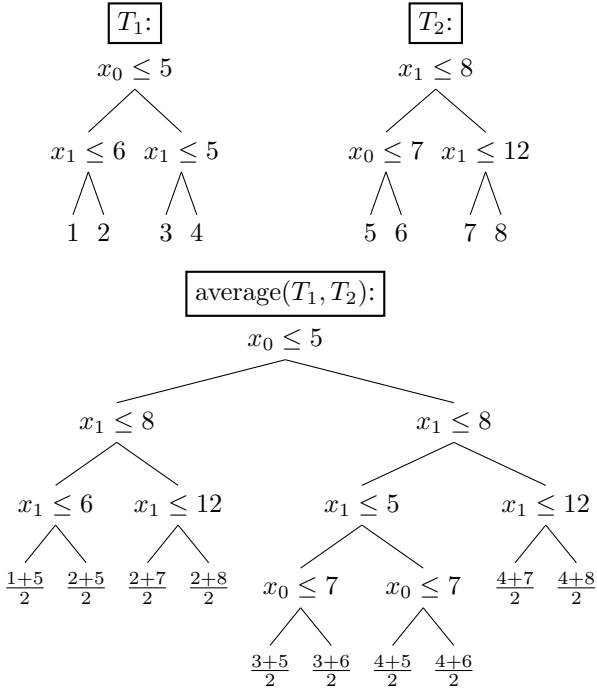
\begin{figure}
  \begin{tikzpicture} [sibling distance=0mm, label/.style = {draw, rectangle, thick}]
  \tikzset{level distance = 11mm}
    \begin{scope}[shift={(-2cm,0)}]
      \node[label] at (0,0.7) {$T_1$:};
      \Tree
        [.$x_0\leq5$
          [.$x_1\leq6$ $1$ $2$ ]
          [.$x_1\leq5$ $3$ $4$ ]
        ]
    \end{scope}
    \begin{scope}[shift={(2cm,0)}]
      \node[label] at (0,0.7) {$T_2$:};
      \Tree
          [.$x_1\leq8$
            [.$x_0\leq7$ $5$ $6$ ]
            [.$x_1\leq12$ $7$ $8$ ]
          ]
    \end{scope}
    \begin{scope}[shift={(0,-3.6cm)}]
      \node[label] at (0,0.7) {$\mathrm{average}(T_1,T_2)$:};
      \Tree
          [.$x_0\leq5$
            [.$x_1\leq8$
              [.$x_1\leq6$ $\frac{1+5}{2}$ $\frac{2+5}{2}$ ]
              [.$x_1\leq12$ $\frac{2+7}{2}$ $\frac{2+8}{2}$ ]
            ]
            [.$x_1\leq8$
              [.$x_1\leq5$
                [.$x_0\leq7$ $\frac{3+5}{2}$ $\frac{3+6}{2}$ ]
                [.$x_0\leq7$ $\frac{4+5}{2}$ $\frac{4+6}{2}$ ]
              ]
              [.$x_1\leq12$ $\frac{4+7}{2}$ $\frac{4+8}{2}$ ]
            ]
          ]
    \end{scope}
  \end{tikzpicture}
  \caption{\label{fig:treeMerge}An example of tree merge}
\end{figure}

Therefore, we designed an algorithm which merges the two trees by walking
simultaneously both trees form the root to the leaves, and performing on-the-fly
optimizations. The algorithm pseudo-code is shown in
Algorithm~\ref{alg:treeAverage}. An example of input and output of the algorithm
is shown in Figure~\ref{fig:treeMerge}.
By this way, we also tend to keep an original aspect of the regression tree
which is that the top-most nodes carry the most important splits (i.e. splits
that strongly reduce the variance of their inner sets of samples).

\begin{algorithm}
  \begin{algorithmic}[1]
    \Function{avgTrees}{$t'$,$t_1$,$t_2$,$w_1$,$w_2$,$\hrect$}
      \State avgNodes(root(t'),root($t_1$), root($t_2$),$w_1$,$w_2$,$\hrect$)
    \EndFunction
    \Function{avgNodes}{$n'$, $n_1$, $n_2$, $w_1$, $w_2$,$\hrect$}
      \If{isLeaf($n_1$)}
         \If{isLeaf($n_2$)}
            \State $\phi_{n'} = \frac{w_1 \phi_{n_1} + w_2 \phi_{n_2}}{w_1 + w_2}$
         \Else
            \State avgNodes($n'$, $n_2$, $n_1$,$w_2$,$w_1$,$\hrect$)
         \EndIf
      \Else
         \State $d \gets \mathrm{splitDim(n_1)}$
         \State $v \gets \mathrm{splitVal(n_1)}$
         \State $v_m \gets \hrect_{d,m}$
         \State $v_M \gets \hrect_{d,M}$
         \If{$v_M \leq v$}
            \State avgNodes($n'$,$n_2$, LC($n_1$),$w_2$,$w_1$,$\hrect$)
         \ElsIf{$v_m < v$}
            \State avgNodes($n'$,$n_2$, UC($n_1$),$w_2$,$w_1$,$\hrect$)
         \Else
            \State split($n'$) $\gets$ split($n$)
            \State $\hrect_{d,M} \gets v$
            \State avgNodes(LC($n')$,$n_2$, LC($n_1$),$w_2$,$w_1$,$\hrect$)
            \State $\hrect_{d,M} \gets v_M$
            \State $\hrect_{d,m} \gets v$
            \State avgNodes(UC($n')$,$n_2$, UC($n_1$),$w_2$,$w_1$,$\hrect$)
            \State $\hrect_{d,m} \gets v_m$
         \EndIf
      \EndIf
    \EndFunction
  \end{algorithmic}
  \caption{\label{alg:treeAverage} The averaging tree algorithm}
\end{algorithm}

\subsection{Pruning trees}
Although our merging procedure helps to reduce the size of the final trees, the
combination of $M$ trees might still lead to a tree of size $O({|t|}^M)$.
Therefore we developed a pruning algorithm which aims at removing the split
nodes which bring the smallest change to the prediction function. The only nodes
that the algorithms is allowed to remove are nodes that are parent of two leafs.
We define the \emph{loss} $\mathcal{L}$ to the prediction function for a node $n$
concerning a hyperrectangle $\hrect_n$ as:
\begin{equation}
      \label{eq:pruningLoss}
      \mathcal{L} = \int\limits_{x \in \hrect_{l}}(\phi'(x) - \phi_{l})\mathrm{d}x + \int\limits_{x \in \hrect_{u}}(\phi'(x) - \phi_{u})\mathrm{d}x
\end{equation}
Where $l$ and $u$ are the lowerchild and upperchild of $n$ respectively, and:
\begin{equation}
      \label{eq:averageApproximator}
      \phi' = \frac{\hrectSize{\hrect_u}\phi_u + \hrectSize{\hrect_l} \phi_{l}}{\hrectSize{\hrect}}
\end{equation}
The prediction function $\phi'$ given by equation~\ref{eq:averageApproximator}
is a weighted average of the prediction functions of both children weighted by the
size of the space concerned by each one. This choice reduces the impact of the
prediction on a leaf when merged with a bigger leaf. Our definition of the loss
$\mathcal{L}$ in equation~\ref{eq:pruningLoss} also considers the size of the
spaces since we compute the integral. The main interest of this method is to
reduce the average error on the whole tree by weighting the cost of an error by
the size of its space.
While most prunning procedures in litterature are centered about reducing the
risk of overfitting, our algorithm (Algorithm~\ref{alg:treePruning}) cares only
about reducing the size of the tree, ensuring that the complexity of the
representation does not go above a given threshold. Since this procedure is not
based on the training set used to grow the forest, it is not necessary to
have an access to the training set in order to prune the tree. When merging the
$M$ trees of a forest, it is crucial to prune the tree resulting of two merge
before applying another merge.

\begin{algorithm}
  \begin{algorithmic}[1]
  \State $\mathrm{splits} = \{ \}$
  \Comment{\parbox[t]{.5\linewidth}{Map from $(\mathrm{node},\mathcal{L})$ to $\phi$,
                                    ordered by $\mathcal{L}$}}
  \ForAll{$n \in \mathrm{preLeafs}(t)$}
    \State $\mathcal{L} = \mathrm{getLoss}(n)$ \Comment{See Eq.~\ref{eq:pruningLoss}}
    \State $\phi = \mathrm{getAverageFunction}(n)$\Comment{See Eq.~\ref{eq:averageApproximator}}
    \State add $((n, \mathcal{L}), \phi))$ to $\mathrm{splits}$
  \EndFor
  \State $\mathrm{nbLeafs} \gets \mathrm{countLeafs}(t)$
  \While{$\mathrm{nbLeafs} > \mathrm{maxLeafs}$}
    \State $((n, \mathcal{L}), \phi)) \gets \mathrm{popFirst}(\mathrm{splits})$
    \State $\phi_n \gets \phi$
    \State $\mathrm{removeChild}(n)$
    \If{$\mathrm{isLastSplit}(\mathrm{father}(n))$}
       \State $n \gets \mathrm{father}(n)$
       \State $\mathcal{L} = \mathrm{getLoss}(n)$ \Comment{See Eq.~\ref{eq:pruningLoss}}
       \State $\phi = \mathrm{getAverageFunction}(n)$\Comment{See Eq.~\ref{eq:averageApproximator}}
       \State add $((n, \mathcal{L}), \phi))$ to $\mathrm{splits}$
    \EndIf
    \State $\mathrm{nbLeafs} \gets \mathrm{nbLeafs} - 1$
  \EndWhile
  \end{algorithmic}
  \caption{\label{alg:treePruning} The tree pruning algorithm}
\end{algorithm}

\section{\label{sec:LearningPolicy}Approximation of the optimal policy}
In this section, we propose three new methods used to choose optimal action for
a given state based on an estimation of the $Q$-value by a regression forest.
While learning of the policy can be computationally demanding since it is
performed offline, it is crucial to obtain descriptions of the policies that
allow very quick computation of the action, given the current state.

\subsection{Learning the continuous policy}
In order to compute the best policy given an approximation of the $Q$-value
$\widehat{Q}$ by a regression forest $F$, we need to solve the following
equation:
\begin{equation}
        \label{eq:bestForestAction}
        \widehat{\pi^*}(s) = \argmax\limits_{a \in A}F((s,a))
\end{equation}
Given $s$, the most straightforward way to compute $\widehat{\pi^*}(s)$
consists in merging all the trees of $F$ projected on $s$ into a single tree
$t'$. Since the size of $t'$
can grow exponentially with the number of trees, we compute an approximation of
$t'$, denoted $\widehat{t'}$ by imposing a limit on the number of leafs using
Algorithm~\ref{alg:treePruning}.  Then it is possible to approximate the best
actions by simply iterating on all the leafs of $\widehat{t'}$ and computing the
maximum of the function $\phi$ of the leaf in its interval. While this solution
does not provide the exact policy which would be induced by $F$, it provides a
roughly good approximation. We refer to this method by
\emph{Fitted $Q$-Iteration}, \emph{FQI} for short.

The FQI is computationally too expensive to be used in online situation:
the computation of a single action requires exploring a potentially large
number of leaves. Therefore, in order to provide a very quick
access to the optimal action for a given state, we propose a new scheme.  By
decomposing the policy function $\pi: S \mapsto A$ into several functions
$\pi_j: S \mapsto A_j$ where $j$ is a dimension of the action space, we can
easily generate samples and use them to train regression forests which provide
estimates of the policy for each dimension. We named this process
\emph{Fitted Policy Forest} and abreviate it by FPF. We use two variants, one
using a piecewise constant model for the nodes, PWC for short, and another
using piecewise linear model for the nodes, PWL for short. We refer to these two
methods by \emph{FPF:PWC} and \emph{FPF:PWL} respectively. Policies resulting of
the FPF algorithm provides a quick access. If such a policy is composed of $M$
trees with a maximal number of nodes $n$, the complexity of getting the action
is $O(M \mathrm{log}(n))$. Since the values used for $M$ does not need to be
high to provide a good approximation~\cite{Ernst2005}, this complexity makes
FPF perfectly suited for real-time applications where online computational
ressources are very limited, such as robotics.

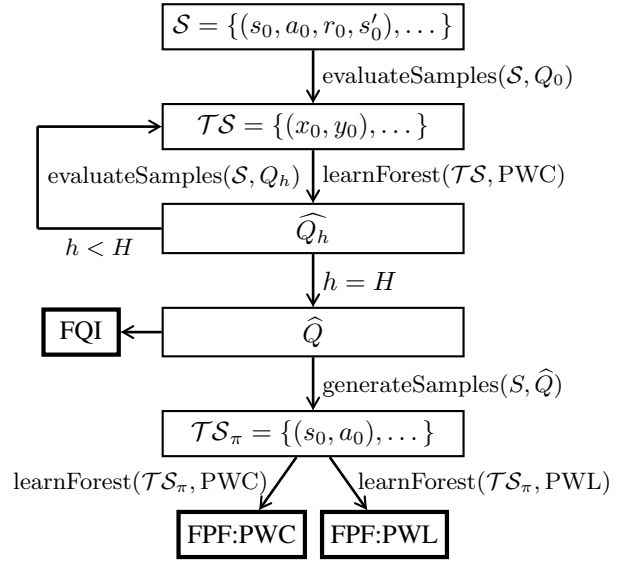
\begin{figure}
  \centering
  \begin{tikzpicture}
  [
    node distance = 7mm,
    block/.style = {draw, rectangle, minimum height=6mm, minimum width=40mm,
                    align=center, thick},
    method/.style = {draw, rectangle, minimum height=6mm, minimum width=10mm,
                     align=center, ultra thick},
    vecArrow/.style={
        decoration={markings,mark=at position 1 with {\arrow[scale=1,thick]{angle 60}}},
        preaction = {decorate},
        postaction = {draw,line width=1pt, shorten >= 1pt}},
    line/.style={draw, line width=1pt}
  ]
  \node[block] (Samples)                   {$\mathcal{S} = \{(s_0,a_0,r_0,s'_0), \dots \}$};
  \node[block] (TS)     [below=of Samples] {$\mathcal{TS} = \{(x_0,y_0) , \dots \}$};
  \draw[vecArrow] (Samples) -- node [right] {\small$\mathrm{evaluateSamples}(\mathcal{S}, Q_0)$} (TS);
  \node[block] (QValue) [below=of TS]      {$\widehat{Q_h}$};
  \draw[vecArrow] (TS) -- node [right] {\small$\mathrm{learnForest}(\mathcal{TS}, \mathrm{PWC})$} (QValue);
  \coordinate[left=of QValue, xshift=-6ex] (wp1);
  \coordinate[left=of TS    , xshift=-6ex] (wp2);
  \draw[line] (QValue) -- node [below] {\small$h < H$} (wp1);
  \draw[line] (wp1) -- node [right] {\small$\mathrm{evaluateSamples}(\mathcal{S}, Q_h)$} (wp2);
  \draw[vecArrow] (wp2) -- (TS);
  \node[block] (FinalQ) [below=of QValue] {$\widehat{Q}$};
  \draw[vecArrow] (QValue) -- node [right] {$h = H$} (FinalQ);
  \node[method] (FQI) [left=of FinalQ, xshift=1ex] {FQI};
  \draw[vecArrow] (FinalQ) -- (FQI);
  \node[block] (TSPol)     [below=of FinalQ] {$\mathcal{TS}_\pi = \{(s_0,a_0) , \dots \}$};
  \draw[vecArrow] (FinalQ) -- node [right] {\small$\mathrm{generateSamples}(S, \widehat{Q})$} (TSPol);
  \node[method] (FPF-PWC) [below=of TSPol, xshift = -6 ex] {FPF:PWC};
  \draw[vecArrow] (TSPol) -- node [left] {\small$\mathrm{learnForest}(\mathcal{TS_\pi}, \mathrm{PWC})$}(FPF-PWC);
  \node[method] (FPF-PWL) [below=of TSPol, xshift =  6 ex] {FPF:PWL};
  \draw[vecArrow] (TSPol) -- node [right] {\small$\mathrm{learnForest}(\mathcal{TS_\pi}, \mathrm{PWL})$}(FPF-PWL);

  \end{tikzpicture}
  \caption{\label{fig:flowchart}A flowchart of the different methods}
\end{figure}

\section{\label{sec:Exploration}Exploration}

While MRE~\cite{Nouri2009} provide a strong basis to build exploration
algorithm, we found that its performance can be strongly improved by bringing
three modifications. First we change the equation used to compute the knownness,
second we use bagging technic to improve the estimation of the knownness, and
third we modify the rule used for $Q$-value update.

\subsection{Original definition}

\emph{Multi Resolution Exploration}~\cite{Nouri2009} propose a generic algorithm
allowing to balance the exploration and the exploitation of the samples. The
main idea is to build a function $\knownness: S \times A \mapsto [0,1]$ which
estimate the degree of knowledge of a couple $(s,a) \in S \times A$. During the
execution of the algorithm, when action $a$ is taken in state $s$, a point
$p = (s_1, \dots, s_{\dim S}, a_1, \dots, a_{\dim A})$ is inserted in a kd-tree, called
knownness-tree. Then, the knownness value according to a knownness-tree $\kdtree$
at any point $p$ can be computed by using the following equation:
\begin{equation}
      \knownness(p) = \min \left ( 1, \frac{\cardinality{P}}{\nu}
                                      \frac{\frac{1}{\floor*{\sqrt[k]{n k / \nu}}}}
                                           {\norminf{\hrect}}
                           \right ) 
\label{eq:MRE:knownness}
\end{equation}
where $\nu$ is the maximal number of points per leaf,
$k = \dim (S \times A)$, $n$ is the number of points inside the whole tree,
$P = \leafPoints{\leaf{\kdtree,p}}$ and $\hrect = \leafSpace{\leaf{\kdtree,p}}$.
A crucial point of this equation is the fact that the knownness value depends on
three main aspects: the size of the cell, the number of points inside the cell
and the number of points inside the whole tree. Therefore, if the ratio between
the number of points contained in a cell and its size does not evolve, its
knownness value will decrease.

The insertion of points inside the kd-tree follows this rule: if adding the
point to its corresponding leaf $\l_0$ would lead to a number of points greater
than $\nu$, then the leaf is splitted into two leafs $l_1$ and $l_2$ of the
same size, and the dimension is chosen using a round-robin. Then the points
stored in $l_0$ are attributed to $l_1$ and $l_2$ depending on their value.

MRE also changes the update rule by using an optimistic rule which replace
equation~\eqref{eq:FQI:output} by equation~\eqref{eq:MRE:output}:
\begin{equation}
       y' = \knownness(s,a) y + (1 - \knownness(s,a)) \frac{R_\mathrm{max}}{1 - \gamma} 
\label{eq:MRE:output}
\end{equation}
where $R_\mathrm{max}$ is the maximal reward which can be awarded in a single
step and $y$ is the result obtained by equation~\eqref{eq:FQI:output}. This
update can be seen as adding a transition to a fictive state containing only
self-loop and leading to a maximal reward at every step. This new transition
occurs with probability $1 - \knownness(s,a)$.

\subsection{Computation of the knownness value}
Initial definition of the knownness is given at
Equation~\eqref{eq:MRE:knownness}. Since this definition does only depend on the
biggest dimension, we have the following. Consider a leaf $l_0$ with a knownness
$\kdtree_0$, then adding a point can result in creating two new leafs $l_1$ and
$l_2$ with respective knowledge of $k_1$ and $k_2$ with $k_0 > k_1$ and
$k_0 > k_2$. This leads to the unnatural fact that adding a point in the middle
of other points can decrease the knowledge of all these points.

We decide to base our knowledge on the ratio between the density of points
inside the leaf and the density of points. Thus replacing
Equation~\eqref{eq:MRE:knownness} by Equation~\eqref{eq:alt:knownness}:
\begin{equation}
  \knownness(p) = \min
    \left ( 1,
      \frac
        {\frac{\cardinality{\leafPoints{\leaf{\kdtree}{p}}}}{\hrectSize{\leaf{\kdtree}{p}}}}
        {\frac{n}{\hrectSize{S \times A}}}
    \right ) 
  \label{eq:alt:knownness}
\end{equation}
where $n$ is the total number of points inside the tree. This definition leads
to the fact that at anytime, there is at least one leaf with a knownness
equal to 1. It is also easy to see that there is at least one leaf with a
knownness strictly lower than 1, except if all the cells have the same density.

\subsection{From knownness tree to knownness forest}
In order to increase the smoothness of the knownness function, we decided to
aggregate several kd-trees to grow a forest, following the core idea of
extra-trees~\cite{Geurts2006}. However, in order to grow different kd-trees
from the same input, the splitting process needs to be stochastic. Therefore,
we implemented another splitting scheme based on extra-trees.

The new splitting process is as follows: for every dimension, we choose at
uniformous random a split between the first sample and the last sample. Thus,
we ensure that every leaf contains at least one point. Then we use an heuristic
to choose the best split.

Once a knownness forest is grown, it is easy to compute the knownness value by
averaging the result of all the trees.

\subsection{Modification of the $Q$-value update}
The $Q$-value update rule proposed by MRE improve the search speed, however it
has a major drawback. Since it only alters the training set used to grow the
regression forest, it can only use the knownness information on state action
combination which have been tried. Therefore, even if for a state $s$ and an 
action $a$, $\knownness(s,a) \approx 0$, it might have no influence at all.

In order to solve this issue, we decided to avoid the modification of the
training set creation, thus using Equation~\eqref{eq:FQI:output}. In place of
modifying those samples, we simply update the regression forest by applying the
following modificator on every leaf of every tree:
\begin{equation}
        v' = v \knownness(c) + R_\mathrm{max} (1 - \knownness(c))
  \label{eq:alt:update}
\end{equation}
with $c$ the center of the leaf, $v$ the original value and $v'$ the new value.

\section{\label{sec:Experimentations}Experimental results}

We present experimental results under two different learning setup. First,
the results obtained by FPF in a batch reinforcement learning, second,
the performances obtained by combining MRE and FPF for online
learning.

\subsection{Batch reinforcement learning}
We used three benchmark problems classical in RL to evaluate the perfomances of
the FPF algorithms. While all the methods share the same parameters for
computing the $Q$-value forest, we tuned specifically parameters concerning the
approximation of the policy using the $Q$-value forest. We compared our results
with those presented in~\cite{Pazis2009}, however we do not have access to their
numerical data, and rely only on the
graphical representation of these datas.  Thus, the graphical lines shown for
BAS are approximative and drawn thicker than the other to highlight the noise in
measurement. We present the result separately for the three benchmarks while
discussing results specific to a problem as well as global results. On all the
problems, performances of FPF:PWL are better or at least equivalent to those
achieved by BAS in~\cite{Pazis2009}. This is remarkable, because BAS uses a
set of basic functions specifically chosen for each problem, while our method
is generic for all the problems. The computation cost of retrieving actions
once the policy has been calculated appears as negligeable and therefore
confirms that our approach is perfectly suited for high-frequency control in
embedded systems.

\subsubsection{\label{sec:batch:ips}Inverted pendulum stabilization}
The \emph{inverted pendulum stabilization} problem consists of balancing a
pendulum of unknown length and mass by applying a force on the cart it is
attached to. We use the description of the problem given in~\cite{Pazis2009}.
The state space is composed of the angular position of the pendulum $\theta$ and
the angular speed of the pendulum $\dot{\theta}$, the action space is $[-50,50]$
Newtons, an uniform noise in $[-10,10]$ Newtons is added. The goal is to keep
the pendulum perpendicular to the ground and the reward is formulated as
following:
$$R(\theta, \dot{\theta}, f) = - \left((2\theta/\pi)^2 + \left(\dot{\theta}\right)^2 + \left(\frac{f}{50}\right)^2\right)$$
except if $|\theta| > \frac{\pi}{2}$, in this case the reward is $-1000$ and
the state is considered as terminal. We set the discount rate $\gamma$ to
$0.95$. The transitions of the system follow the nonlinear dynamics of the
system described in~\cite{Wang1996}:
$$\ddot{\theta} = \frac{g \mathrm{sin}(\theta) - \alpha m l \left(\dot{\theta}\right)^2 \frac{\mathrm{sin}(2\theta)}{2} - \alpha \mathrm{cos}(\theta) u}{\frac{4l}{3} - \alpha m l \mathrm{cos}^2(\theta)}$$
where $g$ is the constant of gravity $9.8 [m/s^2]$, $m = 2.0 [kg]$ is the mass
of the pendulum, $M = 8.0 [kg]$ is the mass of the cart, $l = 0.5 [m]$ is the
length of the pendulum, $\alpha = \frac{1}{m + M}$ and $u$ is the final (noisy)
action applied. We used a control step of $100 [ms]$ and an integration step of
$1 [ms]$ (using Euler Method).
The reward used in this description of the problem ensure that policies leading
to a smoothness of motion and using low forces to balance the inverted pendulum
are rated higher than others.

\begin{figure}
  \includegraphics[width=\linewidth]{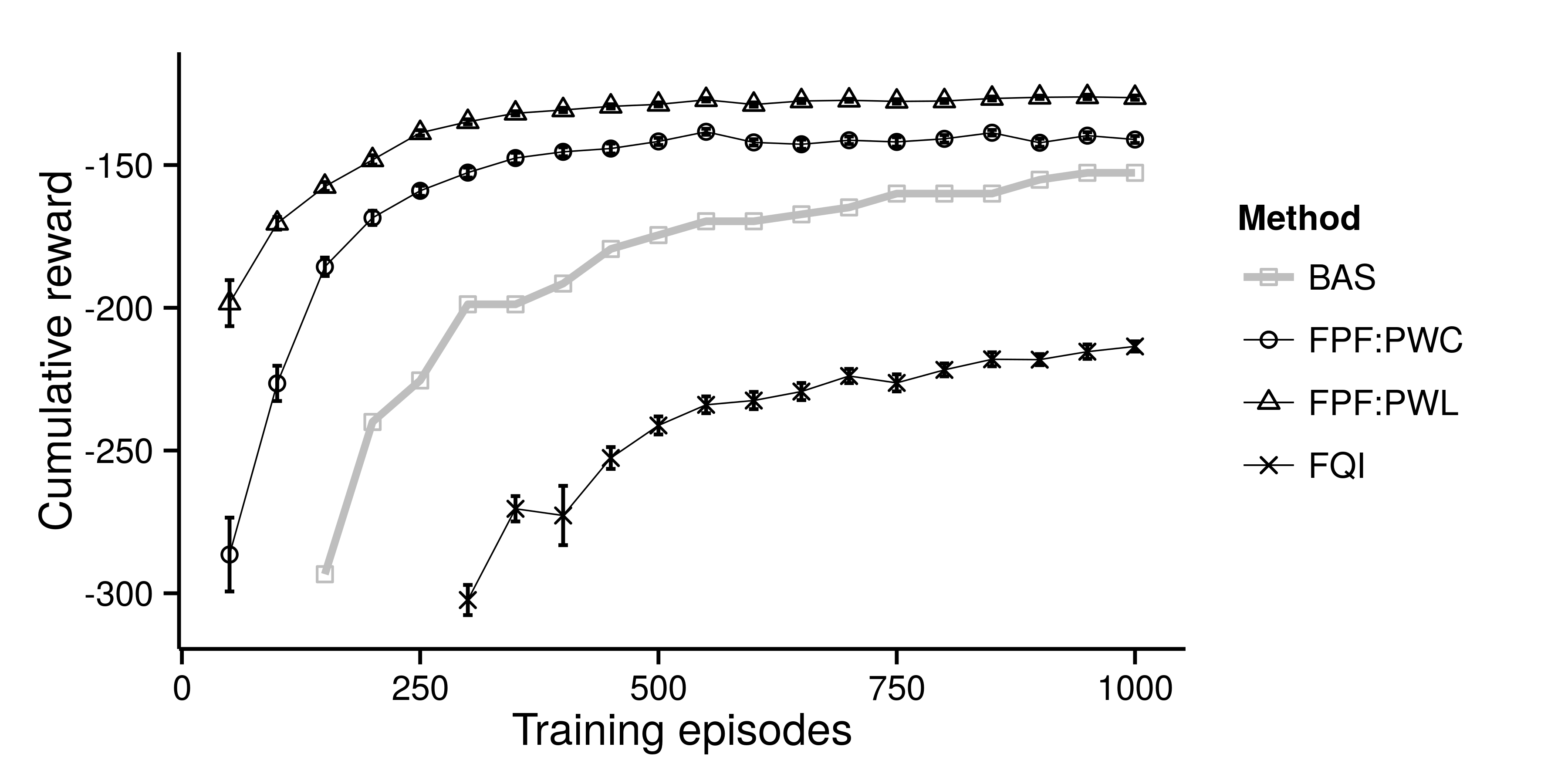}
  \caption{\label{fig:IP:nbEpisodes:algorithm}Performance on the Inverted Pendulum Stabilization problem}
\end{figure}

The training sets were obtained by simulating episodes using a random policy, and
the maximal number of steps for an episode was set to 3000. The performances of
the policies were evaluated by testing them on episodes of a maximal length of
3000 and then computing the cumulative reward. In order to provide an accurate
estimate of the performance of the algorithms, we computed 50 different
policies for each point displayed in Figure~\ref{fig:IP:nbEpisodes:algorithm}
and average their cumulative reward (vertical bars denote 95\% confidence
interval). The parameters used to produce the policies are shown in
Table~\ref{tab:IP:Params}.

Learning a policy from the $Q$-value tree clearly outperform a direct
use on this problem and PWL approximations outperform PWC approximations.
Results for BAS~\cite{Pazis2009} rank systematically lower than both FPF
methods. The huge difference of learning speed between FQI and FPF suggests that
using regression forest to learn the policy from the $Q$-value can lead to
drastical improvements. On such a problem where the optimal policy requires
a fine choice of action, it is not surprising that using linear models to
represent the policy provide higher results than constant models.

The best value for $n_\mathrm{min}$, the minimal number of samples per leaf,
is pretty high (17 for PWC and 125 for PWL). Our understanding of this phenomena
is that the $Q$-value tree tend to slightly overfit the data, additionally, it
uses PWC approximation. Therefore, using it directly lead to an important
quantization noise. Using a large value for $n_\mathrm{min}$ might be seen as
applying a smoothing, which is considered as necessary for regression trees
sampling a stochastic function according to~\cite{Ernst2005}. The need for a
large number of samples is increased for FPF:PWL, because providing an accurate
linear interpolation of a noisy application requires a lot of samples.

\begin{table}
   \caption{\label{tab:IP:Params}Parameters used for Inverted Pendulum Stabilization}
   \vspace{0.2cm}
   \begin{tabular}{l|c|c|c}
      Parameter & FQI & FPF:PWC & FPF:PWL\\
      \hline
      Nb Samples & NA & 10'000 & 10'000\\
      Max Leafs  & 50 & 50     & 50\\
      $k$        & NA & 2      & 2\\
      $n_\mathrm{min}$ & NA & 17 & 125\\
      $M$        & NA & 25 & 25\\
      $V_\mathrm{min}$ & NA & $10^{-4}$ & $10^{-4}$
   \end{tabular}
\end{table}

\subsubsection{Double integrator}
In order to provide a meaningful comparison, we stick to the description of the
problem given in~\cite{Pazis2009} where the control step has been increased from
the original version presented in~\cite{Santamaria1997}. The double integrator
is a linear dynamics system where the aim of the controller is to reduce
negative quadratic costs. The continuous state space consist of the position
$p \in [-1,1]$ and the velocity $v \in [-1,1]$ of a car. The goal is to bring
the car to an equilibrium state at $(p,v) = (0,0)$ by controlling the
acceleration $\alpha \in [-1,1]$ of the car. There are two constraints:
$|p| \leq 1$ and $|v| \leq 1$. In case any of the constraint is violated, a
penalty of 50 is received and the experiment ends. In all other case, the
cost of a state is $p^2 + a^2$. The control step used is $500[\mathrm{ms}]$ and
the integration step is $50[\mathrm{ms}]$, the discount factor was set to
$\gamma = 0.98$.

\begin{figure}
  \includegraphics[width=\linewidth]{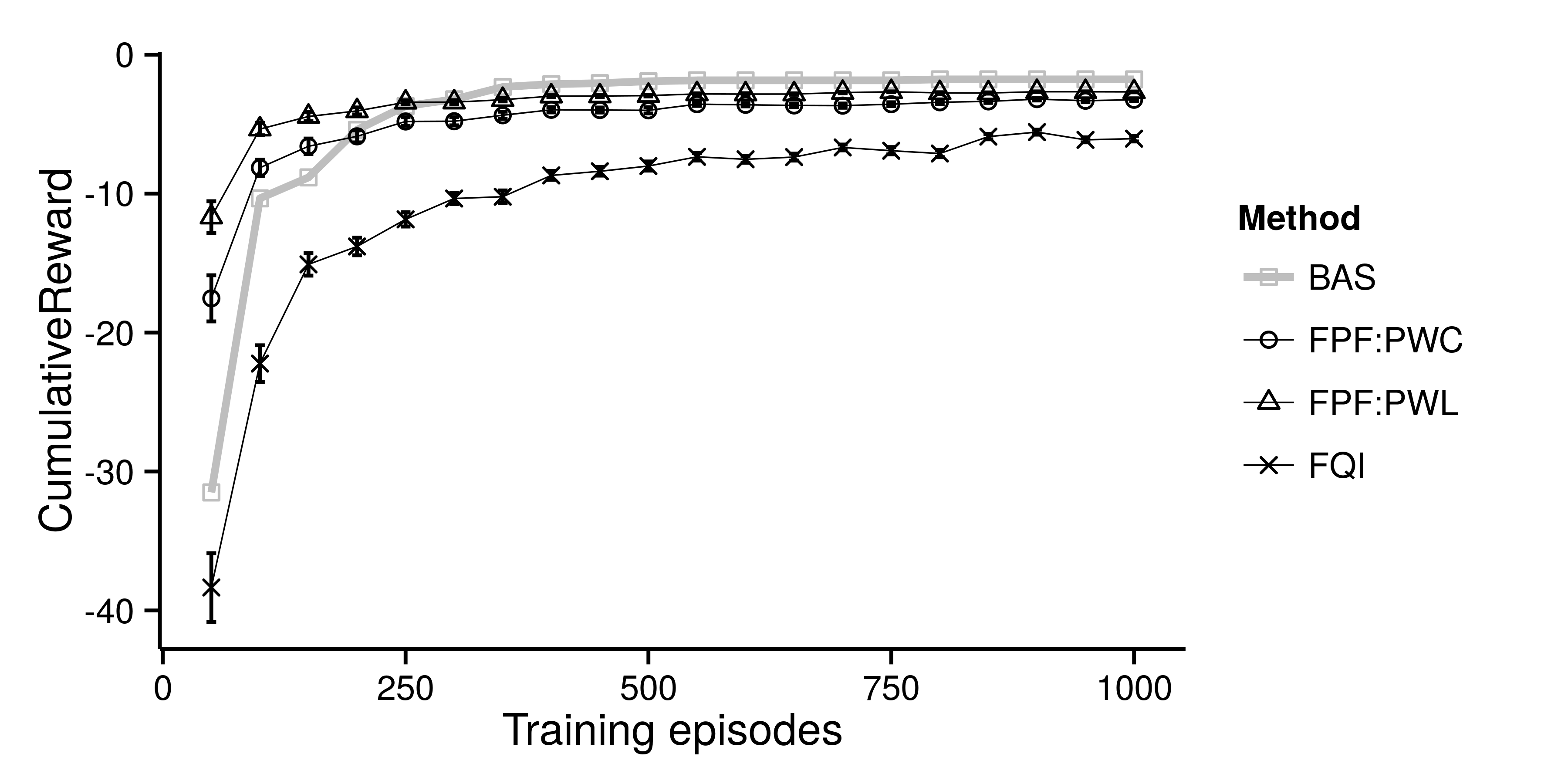}
  \caption{\label{fig:DI:nbEpisodes:algorithm}Performance on the Double Integrator problem}
\end{figure}

The training sets were obtained by simulating episodes using a random policy, and
the maximal number of steps for an episode was set to 200. The performances of
the policies were evaluated by testing them on episodes of a maximal length of
200 and then computing the cumulative reward. In order to provide an accurate
estimate of the performance of the algorithms, we computed 100 different
policies for each point displayed in Figure~\ref{fig:DI:nbEpisodes:algorithm}
and average their results. The parameters used for learning the policy are
shown in Table~\ref{tab:DI:Params}.

On this problem, although none of the proposed methods reach BAS performance
when there are more than 300 learning episodes, FPF:PWL learns quicker than BAS
with a small number of episodes. It is important to note that while our basic
function approximator is constant, a polynome is used for Least-Square Policy
Iteration in~\cite{Pazis2009}, fitting the fact that the optimal policy is
known to be a linear-quadratic regulator~\cite{Santamaria1997}.

\begin{table}
   \caption{\label{tab:DI:Params}Parameters used for Double Integrator}
   \vspace{0.2cm}
   \begin{tabular}{l|c|c|c}
      Parameter & FQI & FPF:PWC & FPF:PWL\\
      \hline
      Nb Samples & NA & 10'000 & 10'000\\
      Max Leafs  & 40 & 40     & 40\\
      $k$        & NA & 2      & 2\\
      $n_\mathrm{min}$ & NA & 100 & 1500\\
      $M$        & NA & 25 & 25\\
      $V_\mathrm{min}$ & NA & $10^{-4}$ & $10^{-4}$
   \end{tabular}
\end{table}

\subsubsection{Car on the hill}
While there has been several definitions of the \emph{Car on the Hill} problem,
we will stick to the version proposed in~\cite{Ernst2005} which was also used as
a benchmark in~\cite{Pazis2009}. In this problem an underactuated car must reach
the top of a hill. The state space is composed of the position $p \in [-1,1]$
and the speed $s \in [-3,3]$ of the car while the action space is the
acceleration of the car $u \in [-4,4]$. If the car violate one of the two
constraints: $p \geq -1$ and $|s| \leq 3$, it receives a negative reward of
$-1$, if it reaches a state where $p> 1$ without breaking any constraint, it
receive a reward of $1$, in all other states, the reward is set to $0$. The car
need to move away from its target first in order to get momentum.

It is well known that the solution to this problem is a bang-bang strategy,
i.e. a nearly optimal strategy exists which uses only the set of actions $\{-4,
4\}$. As stated in~\cite{Pazis2009}, this problem is one of the worst case for
reinforcement learning with continuous action space, since it requires to learn
a binary strategy composed of actions which have not been sampled frequently. It
has been shown in~\cite{Ernst2005} that introducing more actions usually reduce
the performance of the controller. Therefore, we do not hope to reach a
performance comparable to those achieved with a binary choice. This benchmark is
more aimed to assess the performance of our algorithms, in one of the worst case.

While the sample of the two previous algorithms are based on episodes generated
at a starting point, the samples used for the \emph{Car on the hill} problem are
generate by sampling uniformly the state and action spaces. This procedure is
the same which has been used in~\cite{Ernst2005} and~\cite{Pazis2009}, because
it is highly improbable that a random policy could manage to get any positive
reward in this problem. Evaluation is performed by observing the repartition of
the number of steps required to reach the top of the hill from the initial state
$(-0.5, 0)$.

\begin{figure}
  \includegraphics[width=\linewidth]{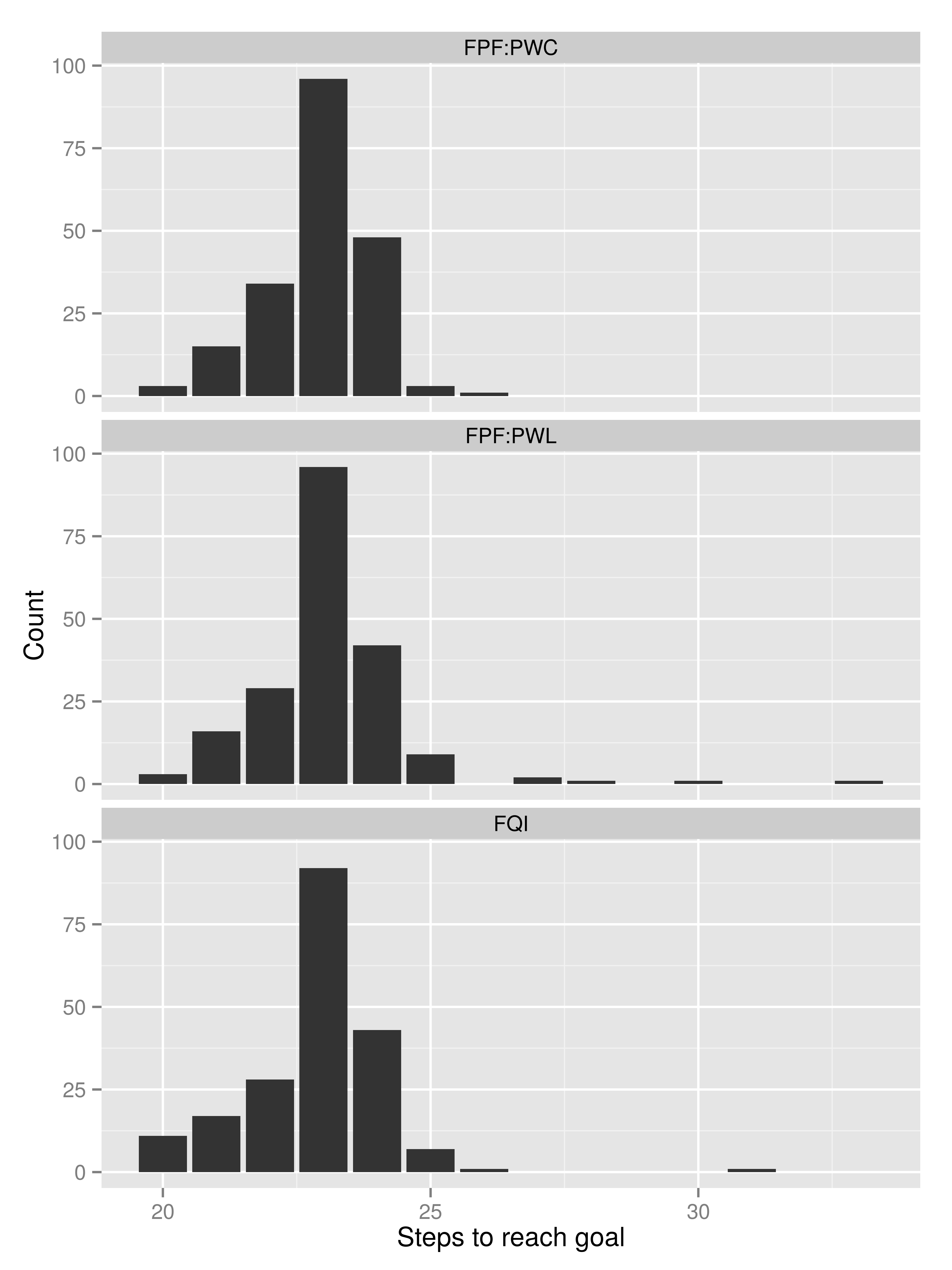}
  \caption{\label{fig:COTH:nbSteps:method}Performance on the Car on the Hill problem}
\end{figure}

We show the histogram of the number of steps required for each method at
Figure~\ref{fig:COTH:nbSteps:method}. For each method, 200 different strategies
were computed and tested. There is no significant difference in the number of
steps required to reach the top of the hill between the different methods. For
each method, at least 95\% of the computed policies led to a number of step in
the interval $[20,25]$. Thus we can consider that an FPF or FQI controller take
20 to 25 steps on average while it is mentioned in~\cite{Pazis2009} that BAS
controller requires 20 to 45 steps on average. Over the six hundred of
experiments gathered across three different methods, the maximal number of steps
measured was 33. Therefore, we can consider that our results strongly
outperforms BAS results.

Car on the Hill is the only problem on which we have not experienced significant
difference between FPF and FQI. Since one of the main advantage of FPF approach
is to reduce the quantization noise of the FQI method, this result is logical.
Although the number of steps required is not reduced by the FPF approach, the
online cost is still reduced by around two orders of magnitude. Therefore, we
can affirm that FPF is highly preferable to FQI on this problem.

\subsubsection{Computational cost}
\begin{figure}
  \includegraphics[width=\linewidth]{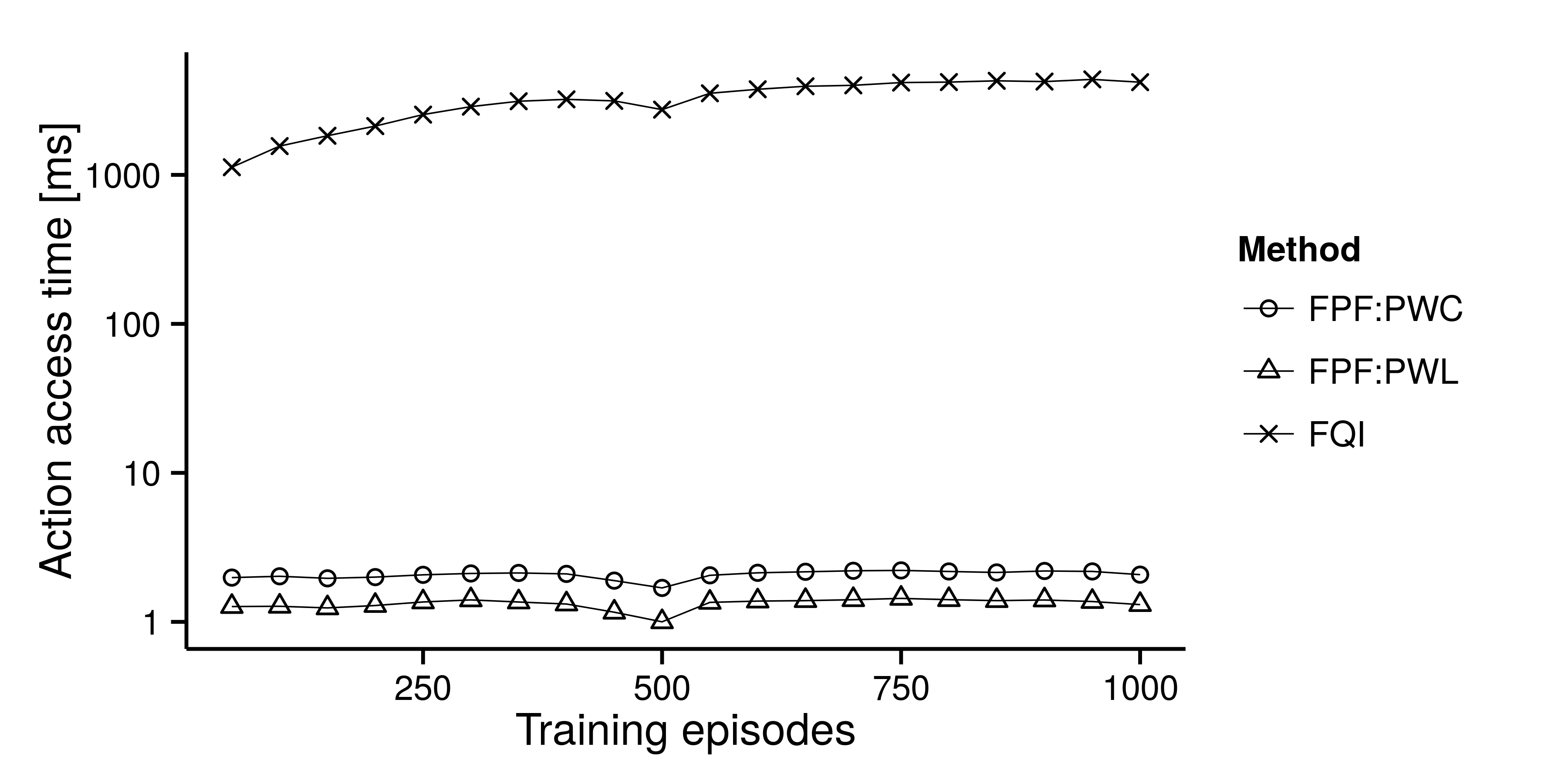}
  \caption{\label{fig:DI:evaluationTime}Evaluation time by episod for the Double Integrator}
\end{figure}

\begin{figure}
  \includegraphics[width=\linewidth]{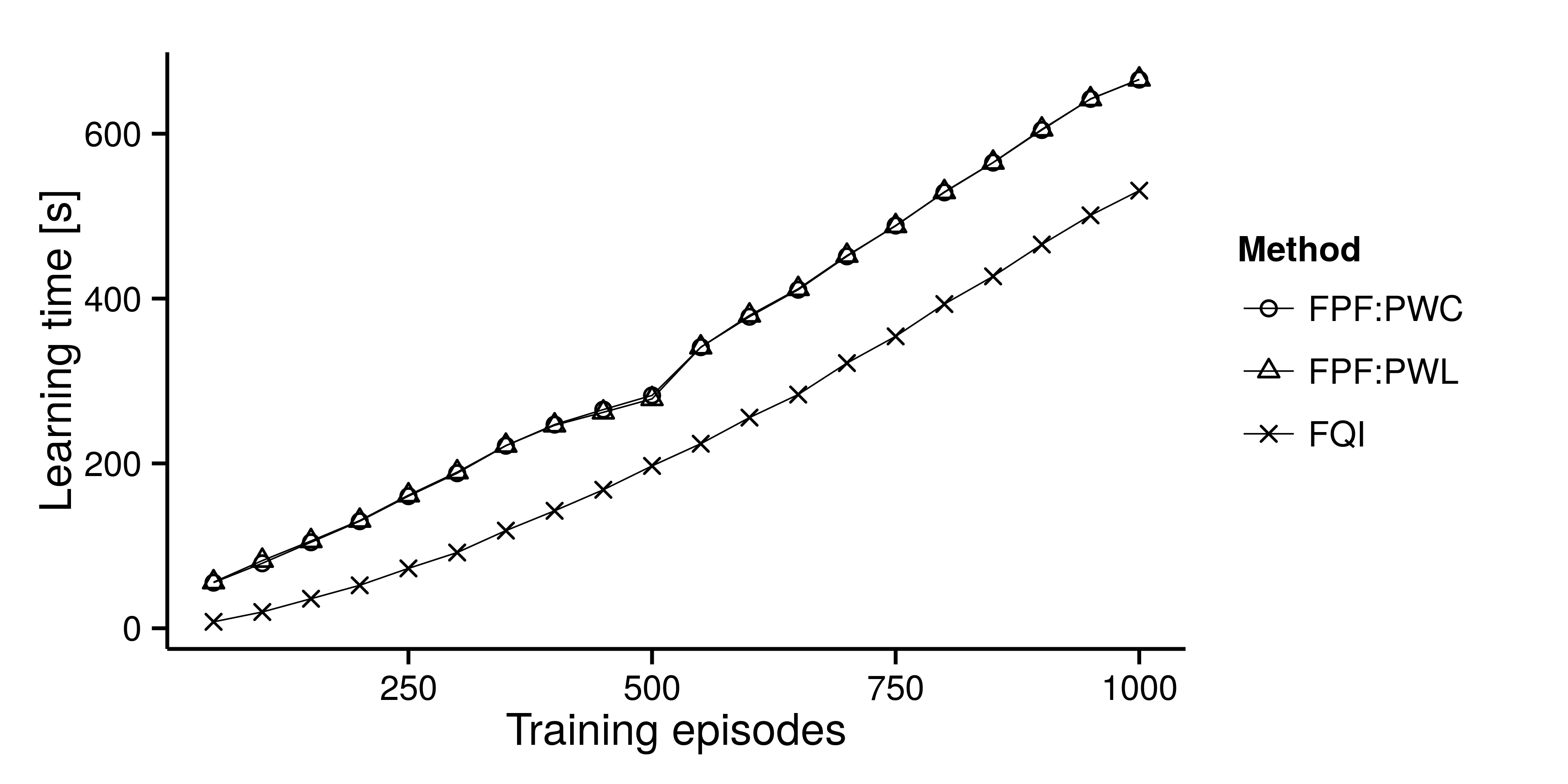}
  \caption{\label{fig:DI:learningTime}Learning time by episod for the Double Integrator}
\end{figure}

As mentioned previously, a quick access to the optimal action for a given state
is crucial for real-time applications. We present the average time spent to
retrieve actions for different methods in Figure~\ref{fig:DI:evaluationTime} and
the average time spent for learning the policies in~\ref{fig:DI:learningTime}.
Experiments were runned using an AMD Opteron(TM) Processor 6276 running at 2.3
GHz with 16 GB of RAM running on Debian 4.2.6.  While the computer running the
experiments had 64 processors, each experiment used only a single core.

We can see that using FPF reduces the average time by more than 2 orders of
magnitude. Moreover, FPF:PWL presents a lower online cost than FPF:PWC, this is
perfectly logical since representing a model using linear approximation instead
of constant approximations requires far less nodes. While the results are only
displayed for the ``Double Integrator'' problem due to the lack of space,
similar results were observed for the two other problems.

It is important to note that the cost displayed in
Figure~\ref{fig:DI:evaluationTime} represents an entire episode simulation, thus
it contains 200 action access and simulation steps.  Therefore, it is safe to
assume that the average time needed to retrieve an action with FPF:PWC or
FPF:PWL is inferior to $50 \mu s$. Even if the CPU used is two orders of
magnitude slower than the one used in the experiment, it is still possible to
include an action access at $200 Hz$.

The additional offline cost of computing the polices required by FPF is
lower than the cost of computing the $Q$-value using FQI when the number
of training episode grows, as presented in Figure~\ref{fig:DI:learningTime}.
Therefore, when it is possible to use FQI, it should also be possible to use FPF
without increasing too much the offline cost.

\subsection{Online reinforcement learning}

We evaluated the performance of the combination of MRE and FPF on two different
problems. First, we present the experimental results on the
\emph{Inverted Pendulum Stabilization} problem and compare them with the
results obtained with random exploration. Second, we exhibit the results on
the \emph{Inverted Pendulum Swing-Up} problem. Since online learning on robots
can be expensive in time and resources, we did not allow for an early phase of
parameter tuning and we used simple rules to set parameters for both problems. In both
problems, the policy is updated at the end of each episode, in order to ensure
that the system is controlled in real-time. In this section, we denote by
\emph{trial} a whole execution of the MRE algorithm on the problem.

\subsubsection{Inverted pendulum stabilization}
This problem is exactly the same as defined in Section~\ref{sec:batch:ips}, but
it is used in a context of online reinforcement learning. The result presented
in this section represent 10 trials of 100 episodes. Each trial was used to
generate 10 different policies, every policy was evaluated by 50 episodes of
3000 steps. Thus, the results concerns a total of 5000 evaluations episodes. 

\begin{figure}
        \centering
  \includegraphics[width=\linewidth]{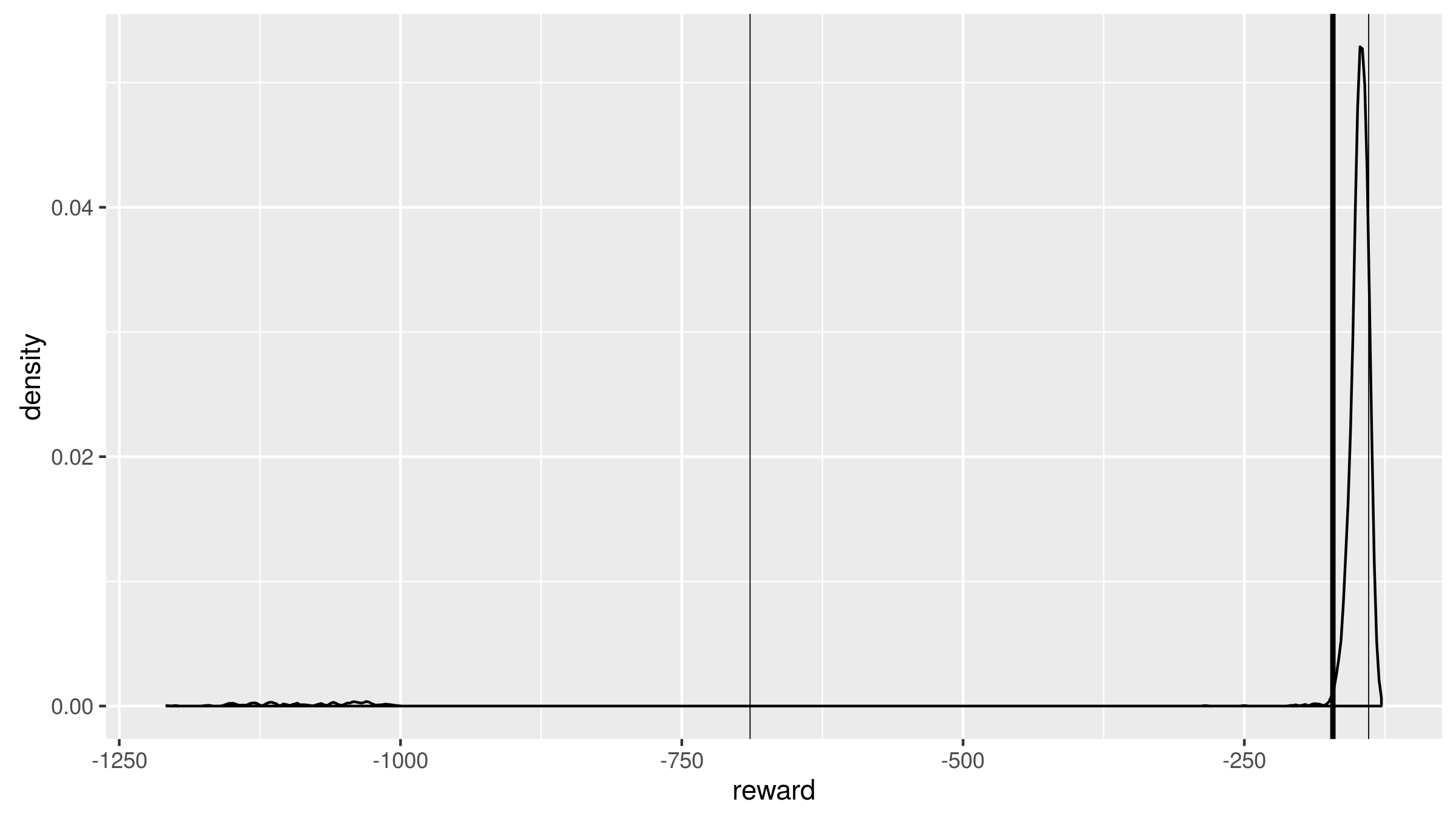}
  \caption{\label{fig:mre:ips:reward}Reward repartition for online learning on
  Inverted Pendulum Stabilization}
\end{figure}

The repartion of reward is presented in Figure~\ref{fig:mre:ips:reward}. The
reward obtained by the best and worst policy are shown as thin vertical lines,
while the average reward is represented by a thick vertical line. Thus, it is
easy to see that there is a huge gap between the best and the worst policy.
Over this 5000 episodes, the average reward per run was $-171$, with a minimum
of $-1207$ and a maximal reward of $-128$. In the batch mode settings,
after the same number of episodes, FPF-PWL obtained an average reward of $-172$,
with a minimal reward of $-234$ and a maximal reward of $-139$. While the
average reward did not significantly improve, the dispersion of reward has
largely increased and in some cases, thus leading to better but also worst
policy. While this might be perceived as a weakness, generating several
policies from the computed $Q$-value is computationally cheap. Then, a few
episodes might be used to select the best policy. From the density of reward
presented in Figure~\ref{fig:mre:ips:reward}, it is obvious that by removing
the worst $10\%$ of the policies, the average reward would greatly improve.

Another point to keep in mind is the fact that the parameters of FPF have not
been optimized for the problem in the MRE setup, while they have been hand-tuned
in the Batch setup. Therefore, reaching a comparable performance without any
parameter tuning is already an improvement.

\subsubsection{Inverted pendulum swing-up}

For this problem, instead of using a mathematical model, we decided to use the
simulator Gazebo\footnote{http://gazebosim.org} and to control it using
ROS\footnote{http://www.ros.org}. Since these two tools are widely accepted in
the robotic community, we believe that exhibiting reinforcement learning
experiments based on them can contribute to the democratization of RL methods
in robotics. We developed a simple model composed of a support and a pendulum
which are bounded by an angular joint. The angular joint is controled in torque
and is underactuated, i.e. the available torque is not sufficient to maintain
the pendulum in an horizontal state. The main parameters are the following: the
mass of the pendulum is $5 [kg]$, the length of the pendulum is $1 [m]$, the
damping coefficient is $0.1 [N m s/\mathrm{rad}]$, the friction
coefficient is $0.1 [N m]$, the maximal torque is
$\tau_{\mathrm{max}} = 15 [N m]$, the maximal angular speed is
$\dot{\theta}_\mathrm{max} = 10 [\mathrm{rad}/s]$ and the control frequency is
$10 [Hz]$. The reward function used is the following
\begin{equation}
r = -\left (
       \norm*{\frac{\theta}{\pi}} +
       \left ( \frac{\tau}{\tau_\mathrm{max}} \right )^2
     \right )
\end{equation}
Where $\theta$ is the angular position of the pendulum (0 denote an upward
position), and $\tau$ represent the torque applied on the axis. If
$\norm*{\dot{\theta}} > \dot{\theta}_{\mathrm{max}}$, a penalty of 50 is applied
and the episode is terminated.

While the system only involves two state dimensions and one action dimension, it
presents two main difficulties: first, random exploration is unlikely to produce
samples where $\theta \approx 0$ and $\dot{\theta} \approx 0$ which is the
target, second, it requires the use of the whole scale of action, large actions
in order to inject energy in the system and fine action in order to stabilize
the system.

The result presented in this section represent 5 trials of 100 episodes. Each
trial was used to generate 10 different policies, every policy was evaluated by
10 episodes of 100 steps. Thus, there is a total of 500 evaluation episodes.

\begin{figure}
  \centering
  \includegraphics[width=\linewidth]{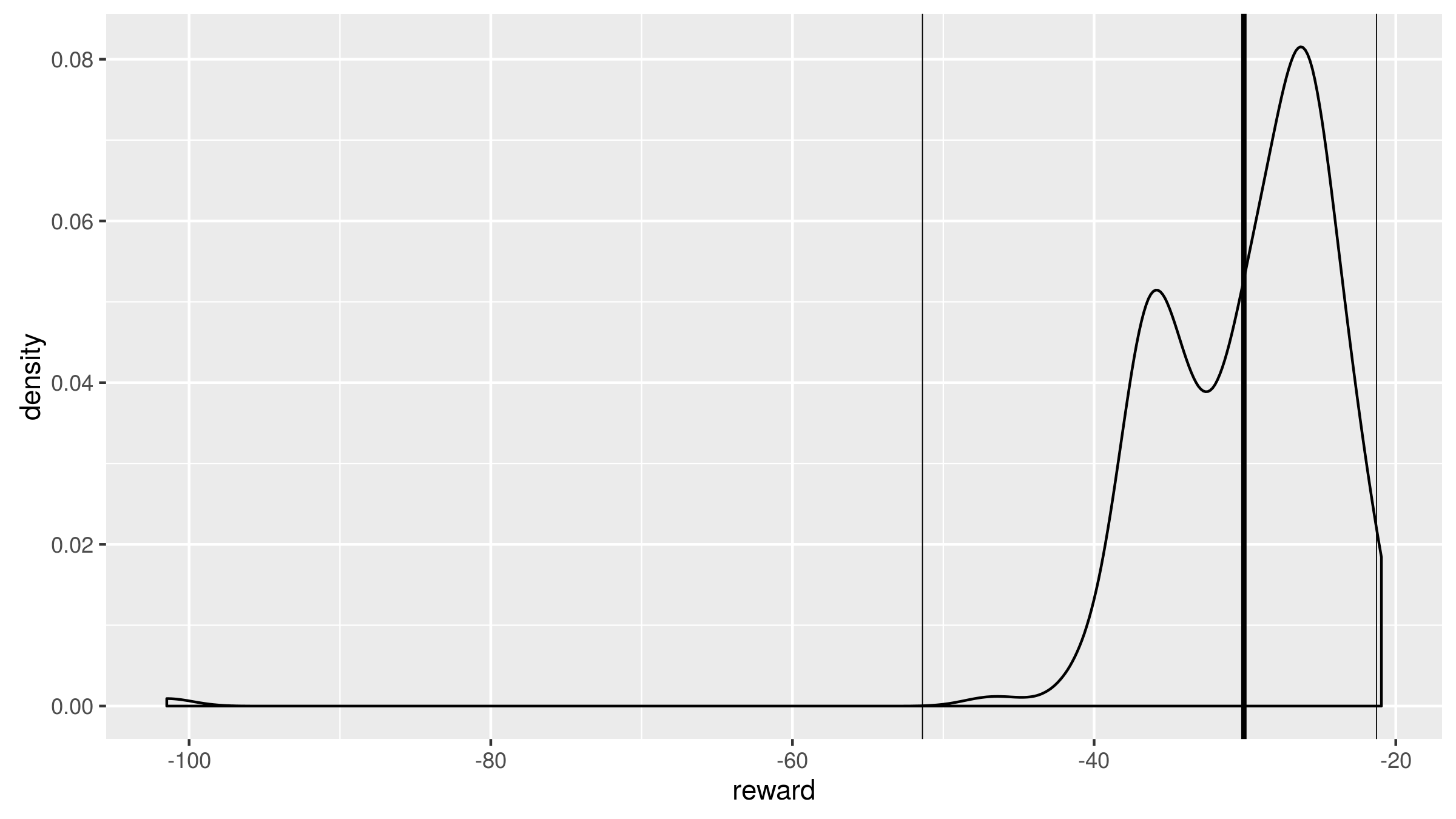}
  \caption{\label{fig:mre:ipsu:reward}Reward repartition for online learning on
  Inverted Pendulum Swing-Up}
\end{figure}

\begin{figure}
  \centering
  \includegraphics[width=.48\linewidth]{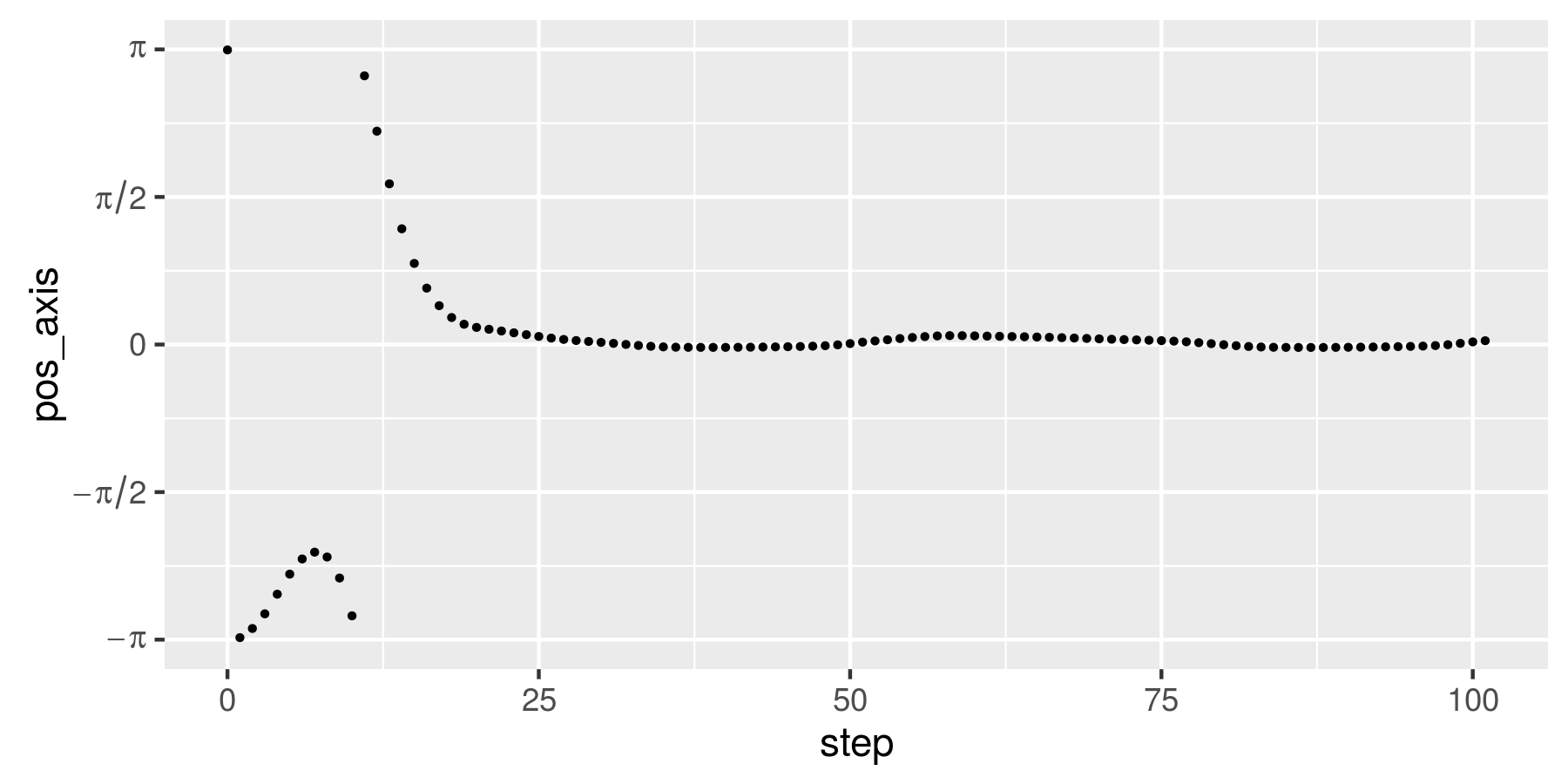}
  \includegraphics[width=.48\linewidth]{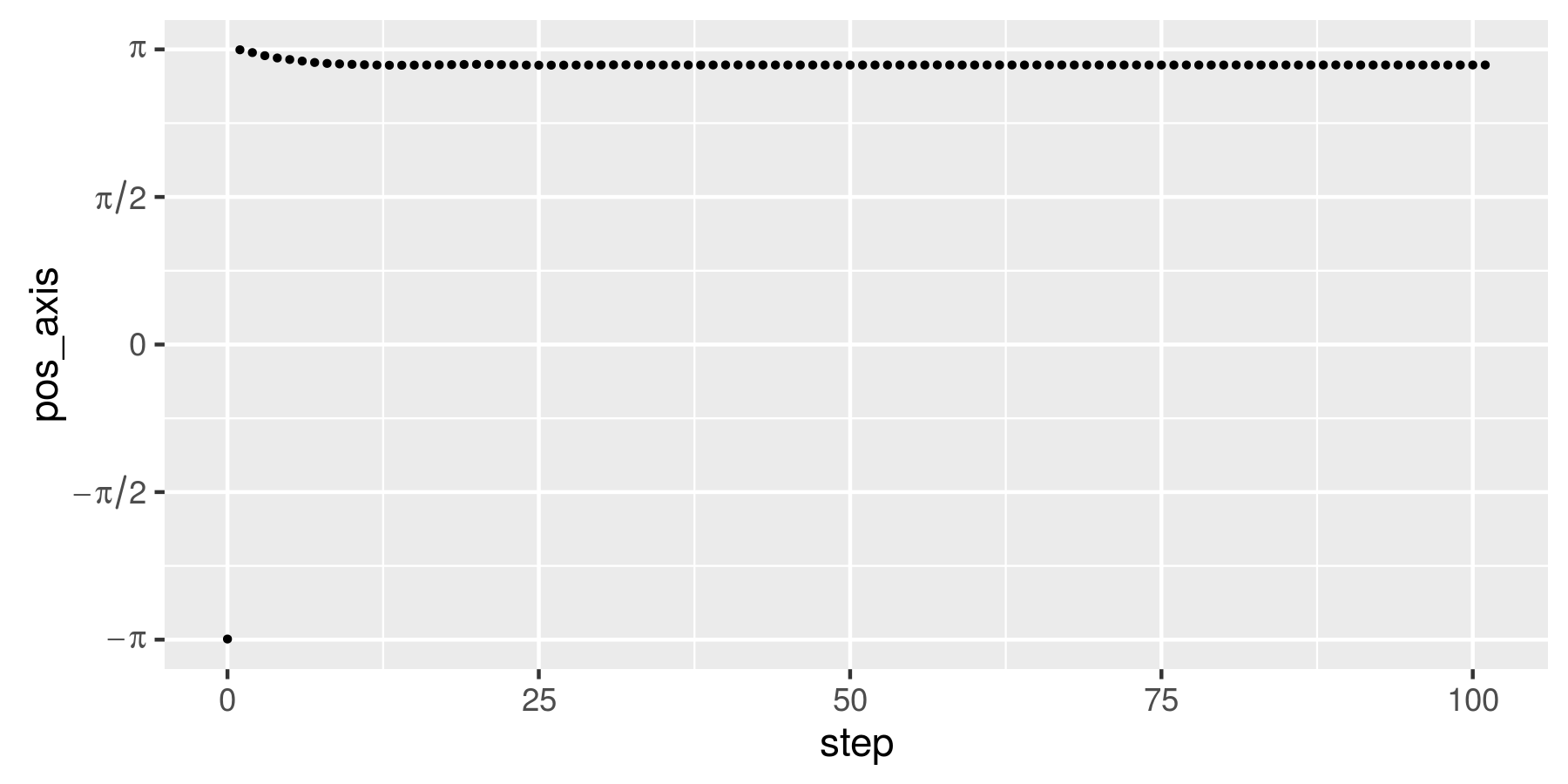}
  \caption{\label{fig:mre:ipsu:best}Best and worst episode for Inverted Pendulum Swing-Up}
\end{figure}

We present the repartition of the reward in Figure~\ref{fig:mre:ipsu:reward}.
The average reward is represented by a thick vertical line and the best and
worst policies rewards are shown by thin vertical lines. Again, we can notice a
large difference between the best and the worst policy. We exhibit the trajectory
of the best and worst evaluation episode in Figure~\ref{fig:mre:ipsu:best}.
While the worst episode has a cumulated reward of $-101$, the worst policy has
an average reward of $-51$. According to the repartition of the reward, we can
expect that very few policies lead to such unsatisfying results, thus ensuring
the reliability of the learning process if multiple policies are generated from
the gathered samples and a few episodes are used to discard the worst policy.

\section{\label{sec:Discussion}Discussion}
Our results show that using FPF does not only allow to drastically reduce the
online computational cost, it also tend to outperforms FQI and BAS, especially
when the transition function is stochastic as in the Inverted Pendulum
Stabilization problem.

Although using piecewise linear function to represent the $Q$-value often leads
to divergence as mentioned in~\cite{Ernst2005}, the same problem did not appear
on any of the three presented problems. In two of the three presented
benchmarks, FPF:PWL yields significantly better results than FPF:PWC and on the
last problem, results were similar between the two method. The possibility of
using PWL approximations for the representation of the policy holds in the fact
that the approximation process is performed only once. Another advantage is the
fact that on two of the problem, the policy function is continuous. However,
even when the optimal policy is bang-bang (Car on the hill), using PWL
approximation for the policy does not decrease the general performance.

Our experiments on the combination of MRE and FPF showed that we can obtain
satisfying results without a parameter-tuning phase. Results also show the
strong variability of the generated policies, thus leading to a natural
strategy of generating multiple policies and selecting the best in a validation
phase.

\section{\label{sec:Conclusion}Conclusion}
This article introduces Fitted Policy Forest, an algorithm extracting a policy
from a regression forest representing the $Q$-value. FPF presents several
advantages: it has an extremely low computational cost to access the optimal
action, it does not require expert knowledge about the problem, it is
particularly successful at solving problems requiring fine actions in stochastic
problems and it can be used with any algorithm producing regression forests.
The effectiveness of our algorithm in a batch setup is demonstrated in three
different benchmarks. The use of FPF in online reinforcement learning is also
discussed and assessed by using MRE as an exploration strategy. Experimental
results suggest that exploration can lead to satisfying results without
requiring any tuning on the parameters. In the future, we also would like to
apply this approach to closed-loop control of Robocup humanoid robots.


\newpage
\bibliographystyle{aaai}
\bibliography{biblio.bib}

\begin{thebibliography}{}

\bibitem[\protect\citeauthoryear{Behnke}{2006}]{Behnke2006}
Behnke, S.
\newblock 2006.
\newblock {Online trajectory generation for omnidirectional biped walking}.
\newblock {\em Proceedings - IEEE International Conference on Robotics and
  Automation} 2006(May):1597--1603.

\bibitem[\protect\citeauthoryear{Breiman}{1996}]{Breiman1996}
Breiman, L.
\newblock 1996.
\newblock {Bagging predictors}.
\newblock {\em Machine Learning} 24(2):123--140.

\bibitem[\protect\citeauthoryear{Busoniu \bgroup et al\mbox.\egroup
  }{2010}]{Busoniu2010}
Busoniu, L.; Babuska, R.; Schutter, B.~D.; Ernst, D.; Busoniu, L.; Babuska, R.;
  Schutter, B.~D.; and Ernst, D.
\newblock 2010.
\newblock {Reinforcement learning and dynamic programming using function
  approximators}.
\newblock  260.

\bibitem[\protect\citeauthoryear{Ernst, Geurts, and Wehenkel}{2005}]{Ernst2005}
Ernst, D.; Geurts, P.; and Wehenkel, L.
\newblock 2005.
\newblock {Tree-Based Batch Mode Reinforcement Learning}.
\newblock {\em Journal of Machine Learning Research} 6(1):503--556.

\bibitem[\protect\citeauthoryear{Geurts, Ernst, and
  Wehenkel}{2006}]{Geurts2006}
Geurts, P.; Ernst, D.; and Wehenkel, L.
\newblock 2006.
\newblock {Extremely randomized trees}.
\newblock {\em Machine Learning} 63(1):3--42.

\bibitem[\protect\citeauthoryear{Koenig and Simmons}{1996}]{Koenig1996}
Koenig, S., and Simmons, R.~G.
\newblock 1996.
\newblock {The effect of representation and knowledge on goal-directed
  exploration with reinforcement-learning algorithms}.
\newblock {\em Machine Learning} 22(1-3):227--250.

\bibitem[\protect\citeauthoryear{Li, Littman, and Mansley}{2009}]{Li2009}
Li, L.; Littman, M.~L.; and Mansley, C.~R.
\newblock 2009.
\newblock {Online exploration in least-squares policy iteration}.
\newblock {\em The 8th International Conference on Autonomous Agents and
  Multiagent Systems}  733--739.

\bibitem[\protect\citeauthoryear{Li, Lue, and Chen}{2000}]{Li2000}
Li, K.-C.; Lue, H.-H.; and Chen, C.-H.
\newblock 2000.
\newblock {Interactive Tree-Structured Regression via Principal Hessian
  Directions}.
\newblock {\em Journal of the American Statistical Association}
  95(450):547--560.

\bibitem[\protect\citeauthoryear{Loh}{2011}]{Loh2011}
Loh, W.-Y.
\newblock 2011.
\newblock {Classification and regression trees}.
\newblock {\em Wiley Interdisciplinary Reviews: Data Mining and Knowledge
  Discovery} 1(1):14--23.

\bibitem[\protect\citeauthoryear{Nouri and Littman}{2009}]{Nouri2009}
Nouri, A., and Littman, M.~L.
\newblock 2009.
\newblock {Multi-resolution Exploration in Continuous Spaces}.
\newblock {\em Advances in Neural Information Processing Systems}  1209--1216.

\bibitem[\protect\citeauthoryear{Pazis and Lagoudakis}{2009}]{Pazis2009}
Pazis, J., and Lagoudakis, M.~G.
\newblock 2009.
\newblock {Binary action search for learning continuous-action control
  policies}.
\newblock {\em Proceedings of the 26th International Conference on Machine
  Learning (ICML)}  793--800.

\bibitem[\protect\citeauthoryear{Peters and Schaal}{2008}]{Peters2008}
Peters, J., and Schaal, S.
\newblock 2008.
\newblock {Reinforcement learning of motor skills with policy gradients}.
\newblock {\em Neural Networks} 21(4):682--697.

\bibitem[\protect\citeauthoryear{Preparata and Shamos}{1985}]{Preparata1985}
Preparata, F.~P., and Shamos, M.~I.
\newblock 1985.
\newblock {\em {Computational geometry: an introduction}}, volume~47.

\bibitem[\protect\citeauthoryear{Puterman}{1994}]{Puterman1994}
Puterman, M.~L.
\newblock 1994.
\newblock {\em {Markov Decision Processes: Discrete Stochastic Dynamic
  Programming}}.

\bibitem[\protect\citeauthoryear{Sanner, Delgado, and de
  Barros}{2012}]{Sanner2012}
Sanner, S.; Delgado, K.~V.; and de~Barros, L.~N.
\newblock 2012.
\newblock {Symbolic Dynamic Programming for Discrete and Continuous State
  MDPs}.
\newblock In {\em Proceedings of the 26th Conference on Artificial
  Intelligence}, volume~2.

\bibitem[\protect\citeauthoryear{Santamaria, Sutton, and
  Ram}{1997}]{Santamaria1997}
Santamaria, J.~C.; Sutton, R.~S.; and Ram, A.
\newblock 1997.
\newblock {Experiments with Reinforcement Learning in Problems with Continuous
  State and Action Spaces}.
\newblock {\em Adaptive Behavior} 6(2):163--217.

\bibitem[\protect\citeauthoryear{Wang, Tanaka, and Griffin}{1996}]{Wang1996}
Wang, H.~O.; Tanaka, K.; and Griffin, M.~F.
\newblock 1996.
\newblock {An approach to fuzzy control of nonlinear systems: Stability and
  design issues}.
\newblock {\em Ieee Transactions on Fuzzy Systems} 4(1):14--23.

\bibitem[\protect\citeauthoryear{Weinstein and Littman}{2013}]{Weinstein2013}
Weinstein, A., and Littman, M.
\newblock 2013.
\newblock {Open-Loop Planning in Large-Scale Stochastic Domains.}
\newblock In {\em 27th AAAI Conference on Artificial Intelligence}, volume~1,
  1436--1442.

\bibitem[\protect\citeauthoryear{Weinstein}{2014}]{Weinstein2014}
Weinstein, A.
\newblock 2014.
\newblock {\em {Local Planning for Continuous Markov Decision Processes}}.
\newblock Ph.D. Dissertation, The State University of New Jersey.

\bibitem[\protect\citeauthoryear{Zamani, Sanner, and Fang}{2012}]{Zamani2012}
Zamani, Z.; Sanner, S.; and Fang, C.
\newblock 2012.
\newblock {Symbolic Dynamic Programming for Continuous State and Action MDPs}.

\end{thebibliography}

\end{document}